\documentclass[runningheads]{llncs}

\providecommand{\authcount}[1]{}

\usepackage{eccv}

\usepackage{eccvabbrv}

\usepackage{url}
\usepackage[utf8]{inputenc}
\usepackage[T1]{fontenc}
\usepackage{url}
\usepackage{booktabs}
\usepackage{amsfonts}
\usepackage{nicefrac}
\usepackage{microtype}
\usepackage{xcolor}
\usepackage{makecell} 
\usepackage{graphicx} 
\definecolor{ourscolor}{HTML}{c2d1e5}
\usepackage{colortbl}
\usepackage{threeparttable}
\usepackage{gensymb}
\usepackage{enumitem}
\definecolor{dgreen}{RGB}{20,139,101}
\usepackage{wrapfig}
\usepackage{lipsum}
\usepackage{tcolorbox}
\usepackage{fontawesome}
\usepackage{marvosym}
\usepackage{multirow}

\usepackage{bbding}
\usepackage{pifont}
\usepackage[linesnumbered,ruled]{algorithm2e}
\usepackage{threeparttable}
\usepackage{tabularx}
\usepackage{xcolor}
\usepackage{array}
\usepackage{rotating}
\usepackage{marvosym}
\usepackage{float}
\usepackage{fontawesome}
\usepackage{wrapfig}
\usepackage{lipsum}
\usepackage{amsmath}
\usepackage{titletoc}
\usepackage{textcomp}
\usepackage{tikz}
\usepackage{subcaption}
\usepackage[table]{xcolor}
\usepackage{booktabs}
\usepackage{adjustbox}
\usepackage{tocloft}
\usepackage{appendix}
\usepackage{listings}
\usepackage{tikz}
\usetikzlibrary{tikzmark, decorations.pathreplacing}
\usepackage{overpic}

\definecolor{dgreen}{RGB}{20,139,101}
\definecolor{front-color}{HTML}{F5FFFA}
\newcommand{\resultone}[1]{{\setlength{\fboxsep}{1.1pt}\colorbox{green!15}{#1}}}
\newcommand{\resulttwo}[1]{{\setlength{\fboxsep}{1.1pt}\colorbox{cyan!15}{#1}}}
\newcommand{\resultthird}[1]{{\setlength{\fboxsep}{1.1pt}\colorbox{yellow!15}{#1}}}
\newcolumntype{P}[1]{>{\centering\arraybackslash}p{#1}}
\newcolumntype{M}[1]{>{\centering\arraybackslash}m{#1}}

\newcommand{\audio}{\includegraphics[height=0.9em]{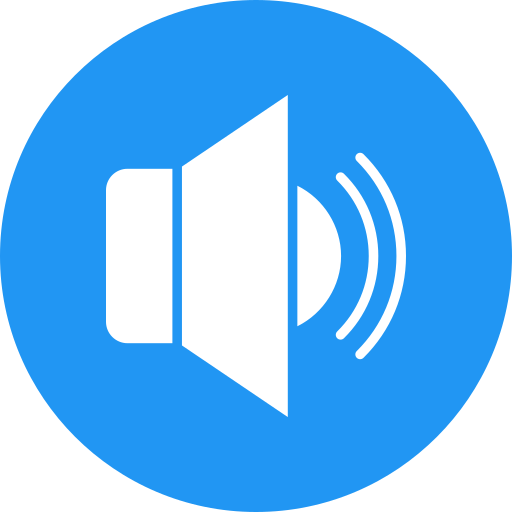}}
\newcommand{\video}{\includegraphics[height=0.9em]{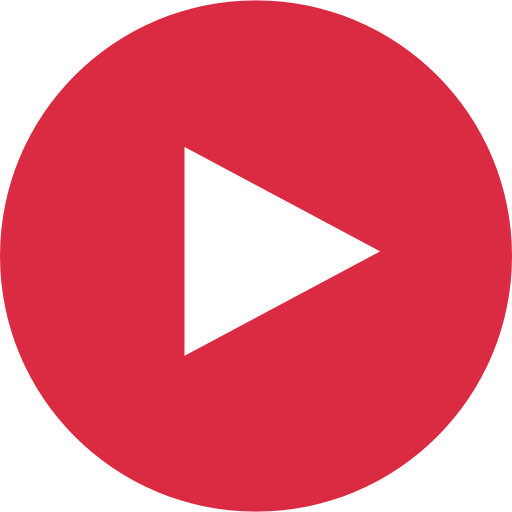}}

\usepackage[accsupp]{axessibility}

\usepackage[pagebackref,breaklinks,colorlinks,linkcolor=eccvblue,citecolor=eccvblue,urlcolor=eccvblue]{hyperref}

\usepackage{orcidlink}

\usepackage{xcolor} 

\usepackage{tcolorbox}

\tcbuselibrary{skins} 

\newcommand{\Preliminary}[2]{
    \vspace{-0.1cm}
    \begin{tcolorbox}[
        colback=white!90!gray,
        colframe=teal!60!black,
        arc=5pt,
        boxsep=5pt,
        left=10pt,
        right=10pt,
        top=2pt,
        bottom=2pt,
        boxrule=0.8pt,
        drop shadow=gray!50!white,
        enhanced jigsaw
    ]
    \vspace{-0.1cm}
        \paragraph{\textbf{\textit{Preliminary #1:}}} #2
    \vspace{-0.1cm}
    \end{tcolorbox}
    \vspace{-0.3cm}
}

\begin{document}

\title{%
\texorpdfstring{%
\begin{tabular}[c]{@{}c@{\hspace{0.2em}}c@{}}
\multirow{2}{*}[0.2em]{\raisebox{2em}{\includegraphics[height=1.8em]{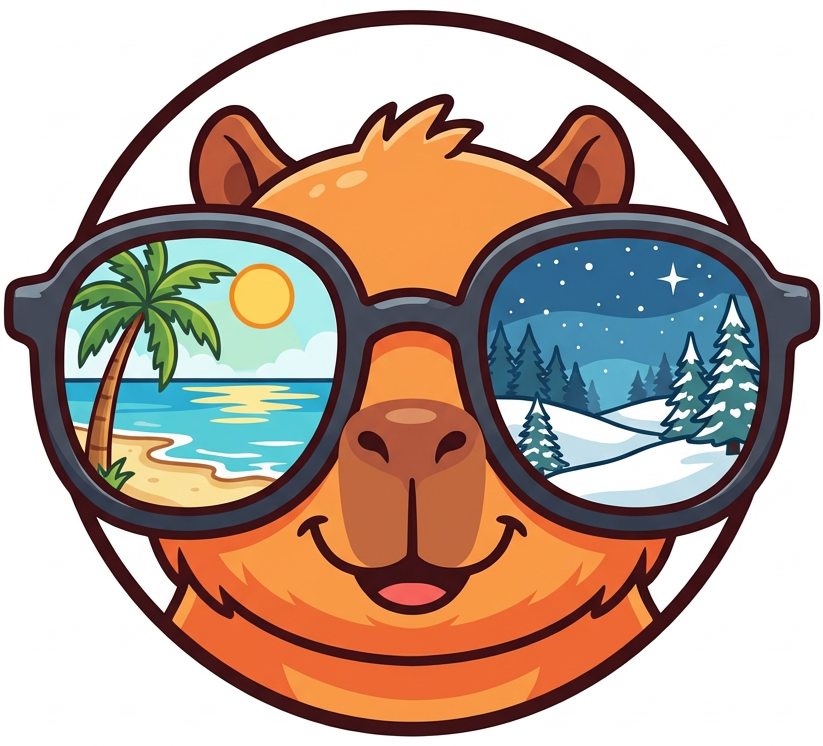}}} &
\begin{minipage}[c]{0.9\linewidth}
\centering
$X$-Stream: Exploring MLLMs as \\ Multiplexers for Multi-Stream Understanding
\end{minipage}
\vspace{-0.5cm}
\end{tabular}%
}{}%
}

\titlerunning{$X$-Stream}

\author{Peiwen Sun\textsuperscript{*}\inst{1}\orcidlink{0009-0005-3016-8554} \and
Xudong Lu\textsuperscript{*}\inst{1}\orcidlink{0009-0007-1699-6286} \and
Huadai Liu\textsuperscript{*}\inst{3}\orcidlink{0009-0004-5782-5641} \and
Yang Bo\inst{2}\orcidlink{0009-0001-0948-9901}\and
Dongming Wu\inst{1}\orcidlink{0000-0003-4938-5813} \and
Huankang Guan\inst{2}\orcidlink{0000-0003-0825-8658} \and
Minghong Cai\inst{1}\orcidlink{0009-0003-7037-4251} \and
Jinpeng Chen\inst{2}\orcidlink{0000-0002-0469-4463} \and
Xintong Guo\inst{2}\orcidlink{0009-0005-5202-5810} \and \\
Shuhan Li\inst{2}\orcidlink{0000-0002-2363-0529} \and
Fang Liu\inst{2}\orcidlink{0000-0002-5763-0172} \and
Rui Liu\inst{2}\orcidlink{0000-0003-2115-8491} \and
Xiangyu Yue\textsuperscript{\tiny \faEnvelopeO}\inst{1}\orcidlink{0000-0002-6887-2046}
}

\authorrunning{P. Sun et al.}

\institute{MMLab, Chinese University of Hong Kong \and
Huawei Research \space \space \inst{3} Hong Kong University of Science and Technology}

\maketitle

\vspace{-1cm}
\begin{figure}[htp]
\centering
    \includegraphics[width=\textwidth]{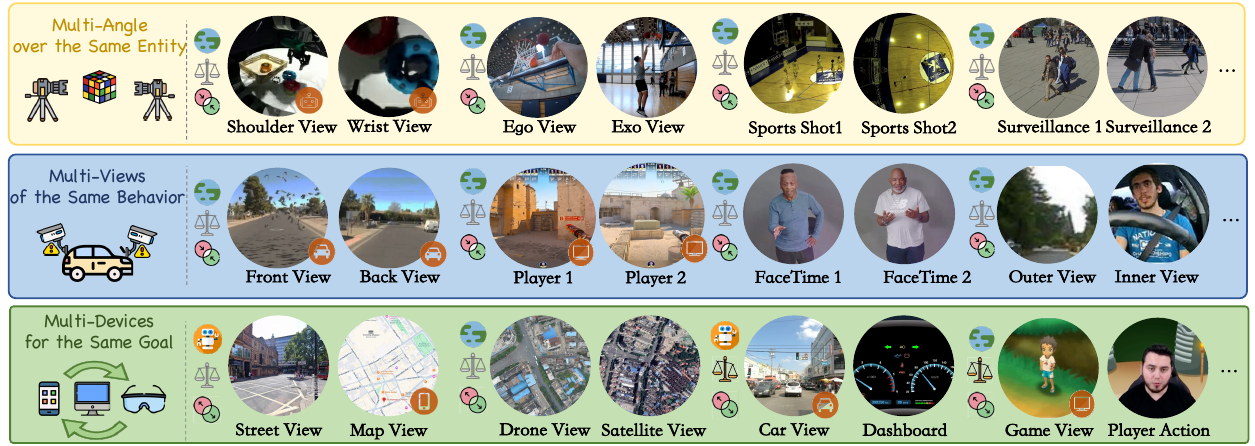}
\vspace{-0.5cm}
\caption{Our $X$-Stream, as the first multi-stream streaming benchmark, encompasses a diverse range of scenarios featuring multi-angle, multi-view, and multi-device capabilities. \raisebox{-0.15em}{\includegraphics[height=0.9em]{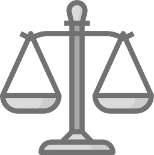}} and \raisebox{-0.15em}{\includegraphics[height=0.9em]{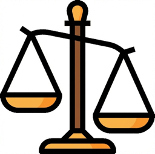}} mean balanced and imbalanced streams. \raisebox{-0.15em}{\includegraphics[height=0.9em]{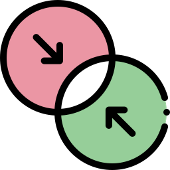}} and \raisebox{-0.15em}{\includegraphics[height=0.9em]{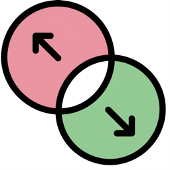}} mean the same domain and different domain streams. \raisebox{-0.15em}{\includegraphics[height=0.9em]{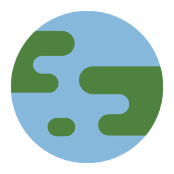}} and \raisebox{-0.15em}{\includegraphics[height=0.9em]{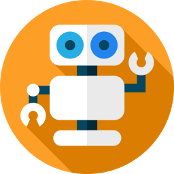}} mean the real-world and synthesized pairs. 
}
\vspace{-1.3cm}
\label{fig:teaser}
\end{figure}

\begin{abstract}

While video streaming understanding has made significant strides, real-world applications, such as live sports broadcasting, autonomous driving, and multi-screen collaboration, inherently demand continuous, multi-stream interactions. 
However, existing benchmarks are confined to single-stream paradigms, leaving a critical gap in evaluating online, cross-stream reasoning. 
To bridge this, we introduce \textbf{$X$-Stream}, the \textbf{first benchmark} dedicated to \textbf{multi-stream} streaming understanding. 
Comprising 4,220 rigorously curated QA pairs across 932 videos, $X$-Stream evaluates 11 subtasks across multi-window, multi-view, and multi-device scenarios. 
Crucially, our dataset is constructed using a novel dual-verification pipeline that prevents over-reliance on a single stream.
Furthermore, we pioneer the conceptualization of multi-modal large language models (MLLMs) as \textbf{naive multiplexers}, systematically evaluating their performance through the lens of Signal Multiplexing Theory. 
Our extensive online inference experiments reveal a stark reality: state-of-the-art MLLMs struggle significantly with concurrent streams, achieving only $\sim$50\% score and exhibiting a poor proactive ability. 
Ultimately, $X$-Stream exposes the \textbf{trade-off} of current multiplexing schemes, providing both a practical evaluation protocol and empirical guidance for next-generation multi-stream agents. 
Code and data are released at \href{https://peiwensun2000.github.io/xstream/}{\faGlobe homepage}.

\vspace{-0.2cm}
\end{abstract}

\vspace{-0.2cm}
\section{Introduction}
\label{sec:intro}
\vspace{-0.3cm}

Propelled by the rapid evolution of Large Language Models (LLMs) like ChatGPT~\cite{singh2025openai}, Gemini~\cite{gemini3pro2025}, and Claude~\cite{anthropic2025claude_sonnet_45_system_card}, AI has successfully transitioned from academic research to everyday application. Following this trajectory, recent models~\cite{qian2025dispider,chen2024videollm} are pushing these boundaries further; by incrementally processing incoming text and frames, they unlock real-time understanding and interactive capabilities for long-form, single-stream streaming videos.

Beyond the challenge of single continuous streams, modern real-world perception increasingly demands multi-stream collaboration. This spans a remarkably wide range of applications, from multi-screen coordination in office environments and orchestrating live feeds in sports broadcasting, to cooperative navigation between mobile maps and smart glasses, and the synchronization of shoulder and wrist cameras on robotic arms. 
For instance, with over 40 distinct broadcast cameras operating at a World Cup game, \textit{“how to automatically select and broadcast the optimal stream during a live football game?”}
The breadth of these scenarios underscores the \textbf{immense practical potential} of multi-stream perception. Consequently, developing such capabilities across multiple simultaneous video streams has become a critical imperative for next-generation AI systems.

\begin{figure}[t]
\centering
    \includegraphics[width=\textwidth]{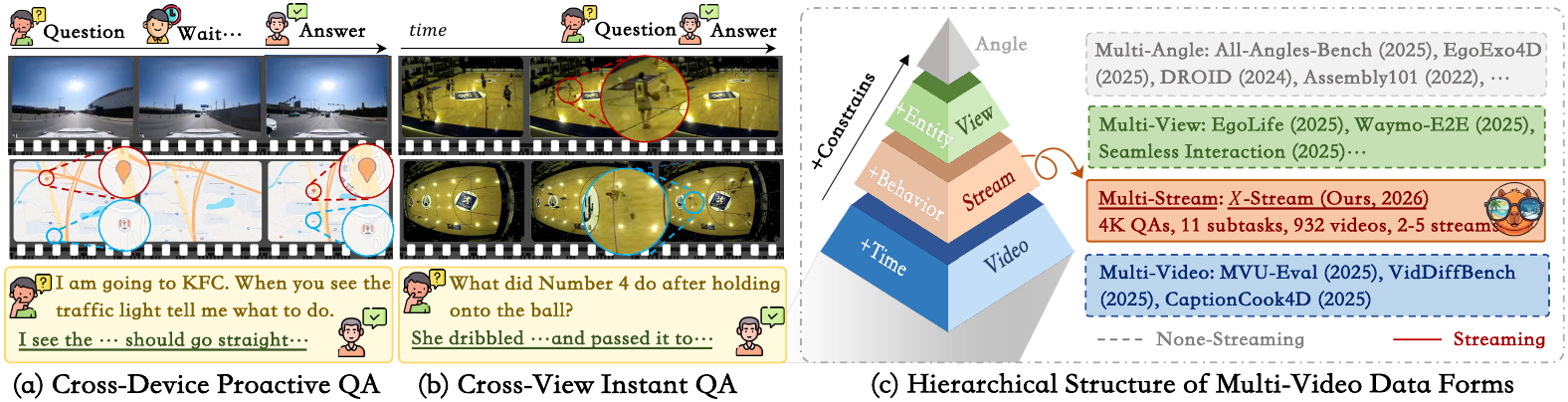}
\vspace{-0.5cm}
\caption{The illustration of the multi-streaming task. Fig.(a) and (b) showcase the practical examples in daily life.
Essentially, the multi-streaming task involves multiple videos with temporal constraints and alignment, requiring the synchronization of video timestamps, as shown in Fig.(c). However, compared to multi-view and multi-angle, it also necessitates important streaming properties to fit the online applications.}
\vspace{-0.5cm}
\label{fig:real_teaser}
\end{figure}

Previous multi-video datasets typically lack streaming characteristics, as well as long-duration, accurately timestamped multi-stream annotations.
During data construction, we identify the over-reliance on a single stream (i.e., single-stream shortcut) as a strong impediment to high-quality data. Then, a novel data protocol and pipeline are used to guarantee the necessity and sufficiency of multi-stream inputs. Finally, as illustrated in Fig.~\ref{fig:real_teaser}(a-b), we introduce \textbf{$X$-Stream} (pronounced ``extreme''), the \textbf{first Multi-Stream Streaming Understanding benchmark}. $X$-Stream comprehensively evaluates models through 4,220 carefully curated QA pairs spanning 932 videos and 451 takes from diverse domains, including daily life, gaming, sports, and autonomous driving. Specifically, the benchmark systematically assesses 4 multi-stream core capabilities across 3 progressive dimensions encompassing 11 sub-tasks: ranging from foundational multimodal perception (e.g., visual/audio/temporal grounding and counting), to high-level logical cognition (e.g., spatial/causal reasoning and anomaly detection), and ultimately to complex decision-making (e.g., behavior planning). Crucially, a higher score across all levels demands the continuous integration of multi-stream omni-modality cues.

In telecommunication, the process of combining multiple signals into one signal over a limited shared medium is called “multiplexing”. Since MLLMs can only handle one token stream at a time, a multiplexer is naturally essential for integrating multiple video streams into one token stream. Therefore, we conceptualize current MLLMs \textbf{as naive multiplexers} processing with a bounded ``bandwidth'' of token processing capacity. To systematically evaluate how models handle concurrent inputs, we develop three distinct multiplexing strategies based on stream division techniques: Spatial, Temporal, and Semantic Division Multiplexing. Finally, we observe the inherent performance trade-offs dictated by these strategies under varying constraints. We reveal that no single approach is universally optimal; rather, their effectiveness is sensitive to the available token bandwidth and the number of concurrent streams. For instance, while spatial division excels in cross-stream referencing, semantic division becomes more necessary to preserve critical information when scaling to three or more streams under tight token budgets. Finally, within this framework, we conduct comparative evaluations of popular models and ablation studies under online streaming inference conditions.

Overall, this paper yields the following key contributions for multi-stream:

\vspace{-0.2cm}
\begin{itemize}[label=\textbf{$\bullet$}]
    \item We propose the \textbf{first multi-stream} streaming benchmark, \textbf{$X$-Stream}, including 4,220 carefully curated diverse QAs spanning 932 videos and 451 takes. Most top-performing models achieve only about 50\% score, and advanced cross-stream skills, like causal reasoning, remain far from application.
    \item To address the over-reliance on a single stream during data construction and evaluation, we introduce a novel data protocol and pipeline that guarantees the \textbf{necessity and sufficiency} of multi-stream inputs. Accordingly, our $X$-Stream benchmark heavily prioritizes the model's multi-stream capabilities.
    \item We also systematically observe the \textbf{inherent trade-offs} introduced by different strategies in multi-stream video multiplexing. Furthermore, we provide a comprehensive analysis to guide future architectural designs.
\end{itemize}
\vspace{-0.5cm}
\section{Related Works}
\label{sec:related}
\vspace{-0.2cm}

\subsection{Multimodal Large Language Models}
\vspace{-0.1cm}

MLLMs have garnered significant attention, driving the emergence of exceptional application-level products. In the realm of video understanding, closed-source models such as GPT-5~\cite{singh2025openai}, Gemini 3 Pro~\cite{gemini3pro2025}, and Doubao-2.0~\cite{seed2026modelcard} currently achieve state-of-the-art performance. Concurrently, open-source models, including InternVL 3.5~\cite{wang2025internvl3}, MiniCPM-V 4.5~\cite{yu2025minicpm}, Qwen 3.5~\cite{qwen3.5}, and DeepSeek-VL2~\cite{wu2024deepseek}, have demonstrated highly competitive capabilities. This rapid progress spans various sub-domains of video analysis, ranging from general comprehension~\cite{fu2025video,yu2019activitynet} to spatial~\cite{sun2025spacevista,wu2025spatial,lang2026longspace} and temporal reasoning~\cite{cheng2025v,chen2024rextime}. Nevertheless, despite these remarkable perceptual breakthroughs, a critical limitation persists: most existing MLLMs are confined to the offline processing of multiple complete videos, lacking the capability to perform online inference on continuous multiple video streams.

\vspace{-0.2cm}
\subsection{Streaming Understanding}
\vspace{-0.2cm}

Streaming requires models to perceive real-time interactions, track forward audio-visual inputs, and respond in the right time. Pioneering efforts \cite{openai_gpt_realtime_docs,defossez2024moshi,wu2023onlinerefer} in the community initially focused on audio streaming capabilities, successfully achieving application-level interaction.

However, streaming video understanding has emerged more recently, primarily constrained by the challenges of processing extensive video tokens. Consequently, advancements are categorized into modeling architectures and data resources.
On the modeling front, research focuses on efficiency and interaction. Frameworks like VideoLLM-online~\cite{chen2024videollm}, StreamingVLM~\cite{xu2025streamingvlm}, and Streamo~\cite{xia2025streaming} optimize memory and processing efficiency for streaming dialogues. Meanwhile, Dispider~\cite{qian2025dispider} addresses perception-reaction conflicts through an asynchronous architecture, and MMDuet2~\cite{wang2025mmduet2} employs multi-turn reinforcement learning to enable autonomous decisions on whether to respond or remain silent.
On the data front, efforts are divided between training resources and evaluation benchmarks. Large-scale training datasets such as HoloAssist~\cite{wang2023holoassist} and EgoBlind~\cite{xiao2025egoblind} facilitate the development of streaming understanding capabilities. For evaluation, StreamingBench~\cite{lin2024streamingbench}, OVO-Bench~\cite{niu2025ovo}, PhoStream~\cite{lu2026phostream}, OmniMMI~\cite{wang2025omnimmi}, and SVBench~\cite{yang2025svbench} assess general streaming understanding and proactive reasoning, while ProactiveVideoQA~\cite{wang2025proactivevideoqa} specifically targets user experience in proactive interaction scenarios. Despite these advancements, existing benchmarks predominantly focus on single stream understanding, leaving a notable gap in specialized evaluations for multi-stream streaming scenarios with open-ended model interactions.

\vspace{-0.4cm}
\subsection{Multi-Video \& Multi-View Understanding}
\vspace{-0.2cm}

From the perspective of video forms, we conceptualize the multi-video data family as a pyramid, illustrated in Fig.~\ref{fig:real_teaser}, organized by increasing constraints: 1) Multi-Video, 2) Multi-Stream, 3) Multi-View, and 4) Multi-Angle.

At the base, \textbf{Multi-Video} Understanding, including MVU-Bench~\cite{peng2025mvu} and video-differencing~\cite{burgess2025video,wu2025vidic} represents the task with the fewest constraints, encompassing multi-video understanding across virtually all scenarios. Further narrowing the scope, \textbf{Multi-View} Understanding, like EgoLife~\cite{yang2025egolife}, Wod-e2e~\cite{xu2025wod}, Seamless-interaction~\cite{agrawal2025seamless}, NuPrompt~\cite{wu2025language} and NuPlanQA~\cite{park2025nuplanqa} mandates multiple perspectives of the same activity—which may involve different subjects—such as front and rear views in autonomous driving or distinct participant feeds in video chats. Finally, at the peak, \textbf{Multi-Angle} Understanding, represented by Assembly101~\cite{sener2022assembly101}, EgoExo4D~\cite{grauman2024ego}, and All-Angle Bench~\cite{yeh2025seeing}, imposes the strictest constraints by necessitating different angles of the same subject at the same time, exemplified by front versus top-down views during assembly tasks or synchronized first-person and third-person perspectives.

In the middle of the hierarchy, \textbf{Multi-Stream} introduces the critical constraint of timestamp alignment. Although Multi-Angle and Multi-View constitute merely a small fraction of the broader Multi-Stream video category in the pyramid, prior approaches \cite{tian2025ego,hasegawa2025promqa} have focused on understanding based on \textbf{entire} multi-view video files rather than evaluating in an \textbf{online streaming} or even real-time manner. Consequently, the field of multi-stream streaming understanding remains unexplored.

\vspace{-0.2cm}
\section{Data Construction}
\label{sec:method}
\vspace{-0.2cm}

In this section, we present a comprehensive construction protocol and statistical analysis of our $X$-Stream benchmark, including data collection in Sec.~\ref{sec:Data_Collection_and_Sources}, task definition in Sec.~\ref{sec:Task_Definition}, annotation pipeline in Sec.~\ref{sec:Data_Pipeline}, and statistical analysis in Sec.~\ref{sec:Benchmark_Statistics}. Further details are available in the appendix.

\vspace{-0.2cm}
\subsection{Data Collection and Sources}
\label{sec:Data_Collection_and_Sources}
\vspace{-0.1cm}

To construct the $X$-Stream benchmark, we systematically gather data across multi-angle, multi-view, and multi-device configurations, as illustrated in Fig.~\ref{fig:teaser}. Our collection comprises 857 hours of raw multi-stream data, featuring 2 to 10 concurrent video streams drawn from over 20 sources. As illustrated in Fig.~\ref{fig:Diversity_analysis}, these sources span eight major domains: driving, sports, robotics, daily routine, chat, surveillance, live streaming, and interface. This raw collection relies on three primary strategies: 1) reformatting metadata from well-established datasets (e.g., Egolife~\cite{yang2025egolife}, Seamless Interaction~\cite{agrawal2025seamless}); 2) combining existing data with simulation techniques to generate multi-device scenarios (e.g., Comma2K-19~\cite{schafer2018commute} with a dashboard); and 3) manually collecting and recording public source data (e.g., Split-screen Game, Map-Street). 
After preprocessing and pairing, we select 160 hours of diverse data across 2-5 streams for further processing.
Due to page limit, further source details and visual previews are available in the appendix.

\vspace{-0.2cm}
\subsection{Task Definition}
\label{sec:Task_Definition}
\vspace{-0.1cm}

Multi-stream streaming understanding demands that a model perceive queries in time, continuously track multiple information streams, and deliver responses at precisely the right moment. Within this dynamic framework, queries are fundamentally categorized by their temporal requirements into instant and forward questions~\cite{lu2026phostream,wang2025proactivevideoqa,lin2024streamingbench}. \textbf{Instant questions} allow the model to generate an immediate response by leveraging retrospective or current context. In contrast, \textbf{forward questions} function as proactive tasks where the necessary conditions for an answer have not yet occurred. For these, the model must actively monitor the incoming streams and wait until the appropriate criteria are met before responding. 
To better evaluate the capabilities of multi-stream models, we systematically model this framework from the dual perspectives of fundamental skills and progressive tasks below.

\begin{figure}[t]
\centering
    \includegraphics[width=\textwidth]{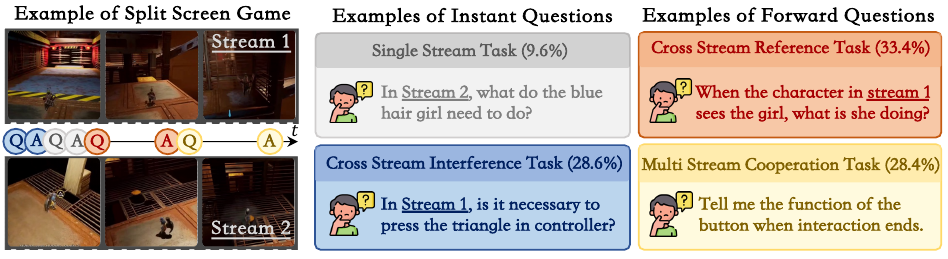}
\vspace{-0.5cm}
\caption{The illustration of the 4 multi-stream abilities. To evaluate these abilities, our $X$-Streaming Benchmark includes 3 progressive dimensions and 11 subtasks.}
\vspace{-0.7cm}
\label{fig:abilities}
\end{figure}

From a skills perspective, this framework encompasses \textbf{4 core capabilities} essential for navigating complex multi-stream environments, as shown in Fig.~\ref{fig:abilities}. Single-stream understanding, as the foundational ability, involves accurately extracting precise information from one specific data stream while operating within a broader multi-stream context. Building upon this, the framework requires cross-stream anti-interference to maintain accuracy by actively filtering out contradictory or irrelevant noise from concurrent streams. Furthermore, it necessitates cross-stream reference alignment to accurately map abstract references in one stream to their corresponding concrete entities or timestamps in another. Finally, the framework culminates in cross-stream cooperation, which demands synthesizing fragmented clues distributed across multiple streams to deduce answers that no single stream could provide alone.

To evaluate these capabilities, the $X$-Stream benchmark is systematically categorized into \textbf{11 progressive tasks}. The foundational level focuses on multimodal perception, encompassing five specific task types: visual, audio, and temporal grounding, counting, as well as saliency
detection. As the complexity increases, the benchmark evaluates high-level logical cognition through five distinct tasks: 3D spatial, causal, counterfactual, and commonsense reasoning, alongside anomaly detection. At the highest level of complexity, the benchmark tests decision-making capabilities, specifically focusing on behavior planning. Crucially, across all three dimensions, generating accurate answers strictly requires the continuous integration of multi-stream cues.

\vspace{-0.3cm}
\subsection{Data Pipeline}
\label{sec:Data_Pipeline}
\vspace{-0.1cm}

\begin{figure}[htbp]
    \centering
    \begin{minipage}[c]{0.3\textwidth}
        \centering
        \includegraphics[width=0.9\textwidth]{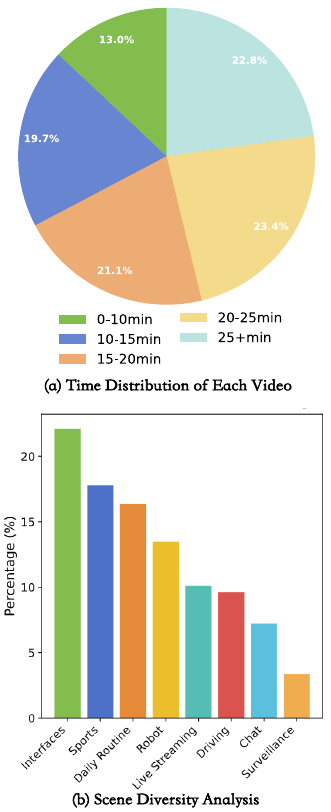}
        \small
        \vspace{-0.1cm}
        \captionof{figure}{Diversity analysis.}
        \label{fig:Diversity_analysis}
    \end{minipage}\hfill
    \begin{minipage}[c]{0.68\textwidth}
        \resizebox{\textwidth}{!}{
            \begin{minipage}{10.5cm} 
                \small
                \begin{algorithm}[H]
                \caption{X-streams Benchmark Pipeline}
                \label{alg:pipeline}
                \KwIn{RawVideo \quad \textbf{Output:} Multi-Stream QA Benchmark}
                \tikzmark{pre-top}
                MultiStreams = Preprocess(RawVideo)\;
                AllCandQA = EmptySet\;
                FinalQA = EmptySet\;
                \tikzmark{pre-bot}
                \tikzmark{qagen-top}
                \For{Video in MultiStreams}{
                    CandQA = GenerateQA(Video)\;
                    Append(AllCandQA, CandQA)\;
                }
                \tikzmark{qagen-bot}
                \For{QA in AllCandQA}{
                    \tikzmark{suf-top}
                    Clip = TrimVideo(MultiStreams, QA.Timestamp)\;
                    \If{Check(Clip, QA.Question) == Correct}{
                        Retain = True\;
                        \tikzmark{suf-bot}
                        \tikzmark{nec-top}
                        \For{SingleStream in Clip}{
                            Ans = Check(SingleStream, QA.Question)\;
                            \If{Ans == Correct}{
                                Retain = False\;
                                \textbf{break}\;
                            }
                        }
                        \tikzmark{nec-bot}
                        \If{Retain == True}{
                            Append(FinalQA, QA)\;
                        }
                    }
                }
                \tikzmark{human-top}
                FinalQA = HumanCheck(FinalQA)\;
                \tikzmark{human-bot}
                \Return FinalQA\;
                \end{algorithm}

                \begin{tikzpicture}[remember picture,overlay]
                  \def\laboff{2mm}
                  \def\amp{4pt}

                  \coordinate (ptop) at (pic cs:pre-top);
                  \coordinate (pbot) at (pic cs:pre-bot);
                  
                  \coordinate (qtop) at (pic cs:qagen-top);
                  \coordinate (qbot) at (pic cs:qagen-bot);
                  
                  \coordinate (stop) at (pic cs:suf-top);
                  \coordinate (sbot) at (pic cs:suf-bot);
                  
                  \coordinate (ntop) at (pic cs:nec-top);
                  \coordinate (nbot) at (pic cs:nec-bot);
                  
                  \coordinate (htop) at (pic cs:human-top);
                  \coordinate (hbot) at (pic cs:human-bot);

                  \coordinate (RefX) at ([xshift=80mm]ptop);

                  \draw[decorate,decoration={brace,amplitude=\amp}]
                    (RefX |- ptop) -- (RefX |- pbot)
                    node[midway,xshift=\laboff,anchor=west,align=center]{Multi-Stream \\ Pre-process};

                  \draw[decorate,decoration={brace,amplitude=\amp}]
                    (RefX |- qtop) -- (RefX |- qbot)
                    node[midway,xshift=\laboff,anchor=west,align=center]{QA \\ Generation};

                  \draw[decorate,decoration={brace,amplitude=\amp}]
                    (RefX |- stop) -- (RefX |- sbot)
                    node[midway,xshift=\laboff,anchor=west,align=center]{Multi-Stream \\ Sufficiency};

                  \draw[decorate,decoration={brace,amplitude=\amp}]
                    (RefX |- ntop) -- (RefX |- nbot)
                    node[midway,xshift=\laboff,anchor=west,align=center]{Multi-Stream \\ Necessity};

                  \draw[decorate,decoration={brace,amplitude=\amp}]
                    (RefX |- htop) -- (RefX |- hbot)
                    node[midway,xshift=\laboff,anchor=west,align=center]{Human \\ Modification};
                \end{tikzpicture}
            \end{minipage}
        }
    \end{minipage}
\vspace{-0.5cm}
\end{figure}

To generate high-quality QA pairs, we follow the pipeline outlined in Alg.~\ref{alg:pipeline}, which consists of four main stages: preprocessing, QA generation, sufficiency and necessity verification, and human verification.

\textbf{Preprocessing.} To establish a baseline video stream for all subsequent generation steps, we first resample the video to 2 FPS. Compared to Gemini’s default 1 FPS, this higher sampling density minimizes the loss of fast-moving actions. Next, we divide and resize the 2 FPS MP4 file into segments smaller than 50MB. This chunking process ensures reliable parallel processing while strictly adhering to storage and transmission limits.

\textbf{QA generation with time stamps.} We employ a hybrid approach: automated generation via MLLMs with rejection sampling, alongside template-based generation for specific subtasks. To construct knowledge- and action-intensive tasks, we randomly sample videos from our dataset and leverage ground-truth metadata, where available, to draft initial questions based on rich curated templates. Finally, we use the Gemini-3-Pro model to refine these template-based questions, rendering them more natural and challenging, and give an accurate answer with a rationale and distractors (if needed). Throughout the construction phase, a balanced type and task distribution is ensured. Additionally, we maintain a roughly 1:1 ratio between multiple-choice and free-form QAs.

\textbf{Shortcut observation.}
During the verification process, we observe that the model tends to generate “pseudo multi-stream” QA pairs that inherently rely on single-stream information. While factually correct, these QAs fail to genuinely evaluate the model's cross-stream capabilities. This degradation primarily manifests in two forms: \textbf{1) Pseudo reference}, where cross-stream anchoring becomes invalid. For example, suppose an ongoing action in Stream 1 (e.g., 'sitting') spans the entire video. In this case, querying it during a momentary event in Stream 2 becomes trivial, making temporal alignment meaningless. \textbf{2) Pseudo cooperation}, which arises from high information redundancy across streams (e.g., overlapping fields of view). In such cases, the model can resolve the query using any single stream, eliminating the need for genuine collaboration or information complementarity.

\textbf{QA verification.} To mitigate the hallucination and the above shortcut, we implement a dual-verification process based on Sufficiency and Necessity. \textbf{Multi-stream Sufficiency} addresses error pairs by ensuring a powerful model can correctly answer the question when given the video clipped to the verified timestamp. Conversely, \textbf{Multi-stream Necessity} eliminates shortcuts by verifying that the model completely fails to answer if provided with only isolated, individual streams. 
During verification, we utilize the videos of high resolution and bit rate, since the target timestamp is already fixed. The final results must satisfy both the necessity and sufficiency criteria.
Together, this approach guarantees high answer accuracy while strictly requiring joint multi-stream understanding.

\textbf{Human verification and modification.} To ensure data quality and safety, we recruit 31 video understanding experts to conduct a rigorous two-round review. During this process, experts verify the clarity of the questions and the accuracy and completeness of the answers, ensuring that responses relied exclusively on audio-visual evidence. Specifically, instant questions had to be fully supported by content preceding the timestamp, whereas forward-looking questions required the timestamp to mark the earliest possible moment the question became answerable, without any premature information leakage. When identifying flawed samples, experts resolve these issues by editing the QA text, correcting task labels, or adjusting timestamps. Conversely, we discard samples that were ambiguous, open to multiple interpretations, insufficiently supported by evidence, or lacking expert consensus. Ultimately, following this meticulous two-stage validation and filtering process, we retain only high-quality samples that were accurate, clear, safe, and strictly grounded. Additionally, the workers' interface, human verification statistics, and error type distribution are provided in the appendix.

\vspace{-0.4cm}
\subsection{Benchmark Statistics}
\label{sec:Benchmark_Statistics}
\vspace{-0.1cm}

\begin{table}[b]
\vspace{-0.5cm}
\captionof{table}{Statistics of multi-domain video/data sources used in our study. “$\sim$'' means “approximately''. Note: one “take” consists of multiple videos, while an individual video may be reused across multiple “takes”.}
\vspace{-0.2cm}
\label{tab:benchmark_comparison}
\centering
\resizebox{\textwidth}{!}{%
\begin{tabular}{llcccccccc}
\toprule
\rowcolor{gray!30} \textbf{Tasks} & \textbf{Dataset} & \textbf{QA} & \textbf{Videos} & \begin{tabular}[c]{@{}c@{}}\textbf{Videos}\\ \textbf{Per Take}\end{tabular} & \textbf{Duration (h)} & \textbf{\begin{tabular}[c]{@{}c@{}}Cross\\ Stream\end{tabular}} & \textbf{\begin{tabular}[c]{@{}c@{}}Open\\ Ended\end{tabular}} & \textbf{Streaming} & \textbf{Proactive} \\ \hline
\multirow{5}{*}{\begin{tabular}[c]{@{}l@{}}Multi-View \&\\ Multi-Video\end{tabular}} 
 & EgoLife-Eval~\cite{yang2025egolife} & 0.3K & 1 & 6 & 20 & \textcolor{red}{\ding{55}} & \textcolor{red}{\ding{55}} & \textcolor{red}{\ding{55}} & \textcolor{red}{\ding{55}} \\
 & ProMQA-Assembly~\cite{hasegawa2025promqa} & 0.4K & 0.2K & 2 & 7 & \textcolor{red}{\ding{55}} & \textcolor{dgreen}{\ding{51}} & \textcolor{red}{\ding{55}} & \textcolor{red}{\ding{55}} \\
 & WaymoQA~\cite{xu2025wod} & 6.4K & $\sim$1K & 2 & $\sim$2 & \textcolor{dgreen}{\ding{51}} & \textcolor{dgreen}{\ding{51}} & \textcolor{red}{\ding{55}} & \textcolor{red}{\ding{55}} \\
 & MVU-Bench~\cite{peng2025mvu} & 1.8K & 5K & 3-5 & 15 & \textcolor{red}{\ding{55}} & \textcolor{red}{\ding{55}} & \textcolor{red}{\ding{55}} & \textcolor{red}{\ding{55}} \\
 & VidDiff~\cite{burgess2025video} & 4.5K & 0.5K & 2 & 3 & \textcolor{red}{\ding{55}} & \textcolor{dgreen}{\ding{51}} & \textcolor{red}{\ding{55}} & \textcolor{red}{\ding{55}} \\ \hline
\multirow{8}{*}{Streaming} 
 & OVO-Bench~\cite{niu2025ovo} & 2.8K & 0.6K & 1 & 85 & \textcolor{red}{\ding{55}} & \textcolor{red}{\ding{55}} & \textcolor{dgreen}{\ding{51}} & \textcolor{dgreen}{\ding{51}} \\
 & StreamingBench~\cite{lin2024streamingbench} & 4.5K & 0.9K & 1 & 136 & \textcolor{red}{\ding{55}} & \textcolor{red}{\ding{55}} & \textcolor{dgreen}{\ding{51}} & \textcolor{dgreen}{\ding{51}} \\
 & Inf-Streams-Eval \cite{xu2025streamingvlm} & 2.5K & 0.5K & 1 & 42 & \textcolor{red}{\ding{55}} & \textcolor{dgreen}{\ding{51}} & \textcolor{dgreen}{\ding{51}} & \textcolor{red}{\ding{55}} \\
 & LiveSports \cite{chen2025livecc} & 1.2K & 0.8K & 1 & 40 & \textcolor{red}{\ding{55}} & \textcolor{dgreen}{\ding{51}} & \textcolor{dgreen}{\ding{51}} & \textcolor{dgreen}{\ding{51}} \\
 & ProactiveVideoQA~\cite{wang2025proactivevideoqa} & 1.4K & 1.4K & 1 & 49 & \textcolor{red}{\ding{55}} & \textcolor{dgreen}{\ding{51}} & \textcolor{dgreen}{\ding{51}} & \textcolor{dgreen}{\ding{51}} \\
 & OmniMMI~\cite{wang2025omnimmi} & 2.3K & 1.1K & 1 & 100 & \textcolor{red}{\ding{55}} & \textcolor{dgreen}{\ding{51}} & \textcolor{dgreen}{\ding{51}} & \textcolor{dgreen}{\ding{51}} \\
 & MMDuet~\cite{wang2025mmduet2} & 2.0K & 2.0K & 1 & 100 & \textcolor{red}{\ding{55}} & \textcolor{dgreen}{\ding{51}} & \textcolor{dgreen}{\ding{51}} & \textcolor{dgreen}{\ding{51}} \\
 & ESTP-Bench \cite{zhang2025eyes} & 2.3K & 1.2K & 1 & 80 & \textcolor{red}{\ding{55}} & \textcolor{dgreen}{\ding{51}} & \textcolor{dgreen}{\ding{51}} & \textcolor{dgreen}{\ding{51}} \\
 & PhoStream~\cite{lu2026phostream} & 5.6K & 0.6K & 1 & 92 & \textcolor{red}{\ding{55}} & \textcolor{dgreen}{\ding{51}} & \textcolor{dgreen}{\ding{51}} & \textcolor{dgreen}{\ding{51}} \\  \hline
\rowcolor{front-color} Multi-Stream 
 & {$X$-Stream (Ours)} & {4.2K} & {0.9K} & 2-5 & {160} & {\textcolor{dgreen}{\ding{51}}} & {\textcolor{dgreen}{\ding{51}}} & {\textcolor{dgreen}{\ding{51}}} & {\textcolor{dgreen}{\ding{51}}} \\ \bottomrule
\end{tabular}
}
\vspace{-0.5cm}
\end{table}

As the pioneering multi-stream streaming benchmark, $X$-Stream is specifically designed to handle complex multi-stream interactions and real-world application scenarios. The benchmark features an accurate dataset comprising 4,220 QAs, 932 videos, and 451 takes. To better fit actual streaming environments, video durations are kept between 5 and 30 minutes, with an average length of 15.8 minutes. While dual-stream videos form the core of the benchmark, approximately 20\% of the data consists of takes with 3 to 5 streams to support more comprehensive multi-stream analysis. Furthermore, around 30\% of the questions incorporate audio or speech information. 
As shown in Tab. \ref{tab:benchmark_comparison}, compared to other benchmarks, ours demonstrates significant advantages in multi-stream and streaming capabilities. Meanwhile, other statistics remain comparable to other popular datasets in QA and video tasks.
Together, these comprehensive metrics demonstrate that $X$-Stream is well-equipped for evaluating advanced, multi-modal streaming applications.

\noindent \textbf{More Information on our $X$-Stream Benchmark}. We encourage readers to consult the appendix for further information, including but not limited to comprehensive source investigations, self-developed data-collection tools, in-depth distribution analyses, rigorous quality control, licensing terms, and QA previews.

\begin{figure}[t]
\vspace{-0.1cm}
\centering
    \includegraphics[width=1\textwidth]{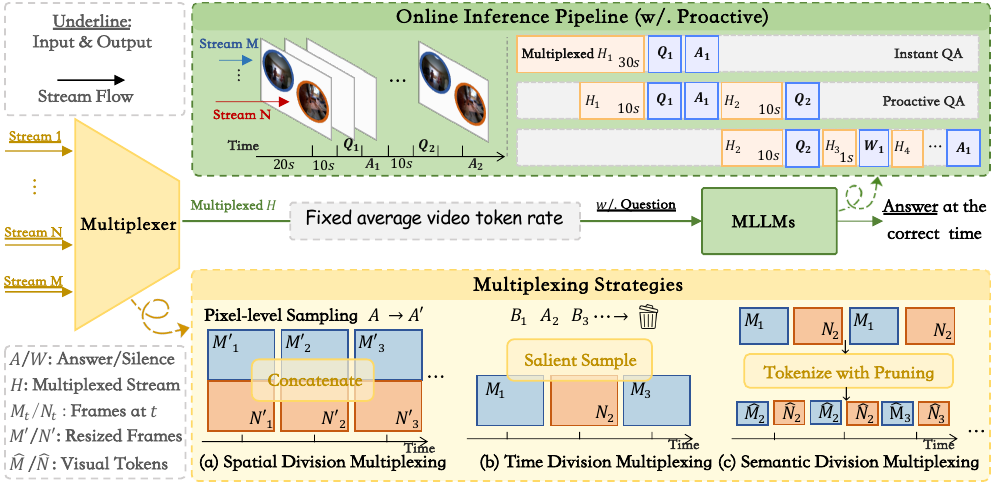}
\vspace{-0.5cm}
\caption{MLLMs can only handle one token stream at a time, making a multiplexer essential for integrating multiple video streams into one token stream. To address this, we investigate three multiplexing strategies and uncover their inherent trade-offs. During evaluation, the model sequentially processes continuous video streams in 1-second intervals while maintaining a sliding memory window for context management.}
\vspace{-0.5cm}
\label{fig:multiplexing}
\end{figure}

\vspace{-0.4cm}
\section{MLLMs as Naive Multiplexers}
\vspace{-0.2cm}

\Preliminary{1}{Multiplexing~\cite{haykin2001communication} is a fundamental method in telecommunication by which multiple signals are combined into one signal over a limited shared medium. 
The method prevents signal interference by dividing them by orthogonality, such as time, frequency, and wavelength.}
\vspace{0.3cm}

Since MLLMs can only take one \textbf{token stream} at a time, integrating multiplexing into multiple \textbf{video streams}  allows us to systematically analyze the approaches to processing multiple streams. 
In practical multi-stream scenarios, MLLMs are inherently constrained by limited context windows and computational budgets. To reflect these real-world limitations, similar to channel bandwidth in telecommunications, a \textbf{limited and fixed average video token rate}, denoted as $C_{max}$, is always enforced. 

We investigate three multiplexing strategies and analyze their inherent trade-offs. The demonstration below illustrates this process using a two-stream setup for clarity, with two concurrent frames $M_t$ and $N_t$ at time $t$.

\noindent\textbf{1) Spatial Division Multiplexing in Fig.~\ref{fig:multiplexing}(a).}
By leveraging the spatial separability of pixels within the frame, this method directly stitches two video streams together and feeds them into MLLMs as a single stream. Formally, we apply a spatial downsampling function $D(\cdot, r)$ with retention ratios $r_m$ and $r_n$. The combined input is constructed by pixel level concatenation as $X_t = \text{Concat}(D(M_t, r_m), D(N_t, r_n))$, subject to the system's token capacity constraint $|\mathcal{T}(X_t)| \leq C_{max}$, where $\mathcal{T}(\cdot)$ denotes the tokenization process. However, this approach requires video re-encoding and audio overlapping prior to input, which introduces extra processing. We also provide grid layout analysis in the appendix.

\noindent\textbf{2) Time Division Multiplexing in Fig.~\ref{fig:multiplexing}(b).}
By processing different video streams as independent inputs, concurrent frames across these streams are assigned identical temporal embedding. Unlike orthogonality mentioned in the preliminary, MLLMs rely on a specific stream identifier (\texttt{<stream N>}) to achieve this kind of separation. Mathematically, we introduce binary indicator variables $\alpha_t, \beta_t \in \{0, 1\}$ to determine whether a frame from stream $M$ or $N$ is sampled at time $t$. This sampling is constrained by the token budget: $\alpha_t |\mathcal{T}(M_t)| + \beta_t |\mathcal{T}(N_t)| \leq C_{max}$. To ensure temporal consistency between $M_t$ and $N_t$, we explicitly align their temporal embeddings. In practice, however, achieving this synchronized temporal encoding is only feasible with open-source models.

\noindent\textbf{3) Semantic Division Multiplexing in Fig.~\ref{fig:multiplexing}(c).} 
Unlike the previous methods that operate on physical dimensions (space and time), this approach multiplexes streams within the semantic space. Mathematically, the basic idea is to formulate a semantic selection function $\mathcal{S}(\mathcal{T}(\cdot), k)$ that retains the $k$ most salient tokens from a given stream, with the constraint $|\mathcal{S}(\mathcal{T}(M_t), k_m)| + |\mathcal{S}(\mathcal{T}(N_t), k_n)| \leq C_{max}$
where $k_m$ and $k_n$ are the allocated token quotas for streams $M$ and $N$. To implement $\mathcal{S}$, following previous training-free token pruning methods~\cite{zhang2025beyond,tangsurge}, we optimally balance token similarity and diversity to maintain salient information with low latency. With the help of extra visual encoders~\cite{radford2021learning,tschannen2025siglip}, a conditional Determinantal Point Process (DPP) kernel matrix is constructed for the candidate tokens as:
{
\setlength{\abovedisplayskip}{2pt}
\setlength{\belowdisplayskip}{2pt}
\begin{equation}
K_{ij} = \text{relevance}_i \cdot \text{similarity}_{ij} \cdot \text{relevance}_j.
\end{equation}
}
where “relevance” denotes how relevant the token is to the current query, and “similarity” represents the degree of similarity between the two tokens.
Then, a greedy Maximum A Posteriori (MAP) inference algorithm is employed to iteratively select the subsets of size $k_m$ and $k_n$. In each stream, the algorithm selects the token with the highest marginal gain and dynamically penalizes the scores of remaining candidates that share high similarity with the selected one. This rigorous selection mechanism ensures that the retained tokens are both highly relevant to the task and visually diverse. Finally, the tokens from different streams are interleaved via time division above.
Since we cannot evaluate proprietary models by the token-level modification, as a workaround, we convert the stream with the most retained tokens back into frames to use as input.

Leveraging these multiplexing schemes, multiple video streams are integrated into a unified token sequence before being input into VLLM for inference.

\vspace{-0.2cm}
\section{Experiments}
\label{sec:experiments}
\vspace{-0.2cm}

In this section, we present a series of experiments on $X$-Stream. We describe the experimental setup, report baseline results, conduct a multiplexing ablation study, and perform a human test to support the LLM-as-a-Judge evaluation.

\vspace{-0.2cm}
\subsection{Experiment Setup}\label{sec:exp_set}
\vspace{-0.1cm}

Following~\cite{lu2026phostream}, we evaluate three categories of baseline models with the Online Inference Pipeline and LLM-as-a-Judge evaluation. We report Instant, Backward, Forward, and comprehensive scores for all models. For further analysis of the Forward setting, we also report the proportions of Early Response (ER, $\downarrow$) and No Response (NR, $\downarrow$). ER denotes any response other than Silent or a placeholder that occurs before Timestamp Proactive. NR denotes that the model produces no response other than Silent or a placeholder within the response window. Then, we use a 2-second response window and run inference for 6 time slots. 
Therefore, achieving a high score on a forward question requires providing the correct answer at the precise moment. However, when calculating the score of multi-stream abilities, we only average the scores of temporally accurate answers to avoid imbalance caused by different answer timings.
Additionally, we cap the average $C_{max}=250$ tokens per video second. However, due to variations in token calculation across models, we employ diverse methods to enforce this limit, such as adjusting playback speed and resizing videos. $r_m$, $r_n$ are dynamically set at the largest value within $C_{max}$. Also, $\alpha_t$, $\beta_t$ are uniformly sampled from $\{0,1\}$.

\begin{table}[bh]
\centering
\vspace{-0.5cm}
\caption{Comprehensive streaming performance comparison of mLLMs on the X-Stream Benchmark (Ours). The symbols ``\video'' and ``\audio'' indicate video and audio support, respectively. Background colors denote the top three results within each scene: \resultone{green (1st)}, \resulttwo{blue (2nd)}, and \resultthird{yellow (3rd)}. Among open-source models, \textbf{bold} and \underline{underlined} highlight the best and second-best.}
\label{tab:main_comparison}
\vspace{-0.2cm}
\resizebox{\textwidth}{!}{%
\begin{tabular}{l|>{\columncolor{gray!15}}c|cccc|cc|cccc}
\toprule
\multirow{2}{*}{\textbf{Model}} &  & \multicolumn{4}{c|}{\textbf{Evaluation Score (\ensuremath{\uparrow})}} & \multicolumn{2}{c|}{\textbf{Forward Time}} & \multicolumn{4}{c}{\textbf{Multi-Stream Abilities}} \\ \cline{3-12} 
 &\multirow{-2}{*}{\textbf{Overall}}& \textbf{Instant} & \textbf{Backward} & \textbf{Forward} & \textbf{Compre.} & \textbf{ER (\ensuremath{\downarrow})} & \textbf{NR (\ensuremath{\downarrow})} & \textbf{\begin{tabular}[c]{@{}c@{}}Single\\ Stream\end{tabular}} & \textbf{\begin{tabular}[c]{@{}c@{}}Multi\\ Coop.\end{tabular}} & \textbf{\begin{tabular}[c]{@{}c@{}}Cross\\ Ref.\end{tabular}} & \textbf{\begin{tabular}[c]{@{}c@{}}Cross\\ Inter.\end{tabular}} \\ \hline
\rowcolor{gray!30} Human Preference \video \space \audio & 91.84 & 91.73 & 95.19 & 85.10 & 97.50 & 9.50 & 2.55 & 94.12 & 92.05 & 90.10 & 98.55 \\ \hline
\multicolumn{12}{l}{\textit{Proprietary Multimodal Models}} \\ \hline
Gemini 3 Pro~\cite{gemini3pro2025} \video \space \audio & \resultone{49.60} & \resultone{73.38} & \resultone{72.23} & \resultone{20.77} & \resultone{82.04} & 73.13 & \resultone{0.23} & \resultone{72.45} & \resultone{71.16} & \resultone{74.79} & \resultone{66.96} \\
GPT-5~\cite{singh2025openai} \video & 27.78 & 44.28 & 37.18 & 6.51 & \resultthird{59.83} & 81.73 & 1.14 & 39.08 & 44.12 & 52.75 & 45.65 \\
GPT-4o~\cite{hurst2024gpt} \video & 22.46 & 37.28 & 32.72 & 4.05 & 47.01 & 87.14 & 0.74 & 34.83 & 34.90 & 43.77 & 37.52 \\
Doubao-Seed-1.8~\cite{seed2026vision} \video & \resulttwo{36.79} & \resultthird{55.49} & \resulttwo{57.18} & \resulttwo{14.52} & 59.13 & 66.19 & 3.95 & 47.55 & 35.69 & 56.52 & \resultthird{60.82} \\ \hline
\multicolumn{12}{l}{\textit{Open-source Multimodal Models}} \\ \hline
Qwen2.5-VL-7B~\cite{Qwen2.5-VL} \video & 25.49 & 40.02 & 36.02 & 8.34 & 45.28 & 68.10 & 11.36 & 43.80 & 41.43 & 42.72 & 40.01 \\
Qwen2.5-Omni-7B~\cite{xu2025qwen25omnitechnicalreport} \video \space \audio & 26.82 & 41.96 & \underline{41.17} & \underline{9.03} & 45.04 & \resulttwo{\underline{53.19}} & 22.51 & 38.60 & 40.80 & 41.86 & 44.35 \\
Qwen3-VL-8B~\cite{bai2025qwen3vltechnicalreport} \video & 26.78 & 43.41 & 33.30 & 7.53 & 51.01 & 78.40 & 6.50 & 49.88 & 43.41 & 33.30 & 51.01 \\
Qwen3-Omni-30B-A3B~\cite{xu2025qwen3omnitechnicalreport} \video \space \audio & \resultthird{\textbf{34.28}} & \resulttwo{\textbf{63.92}} & \resultthird{\textbf{53.40}} & 0.61 & \resulttwo{\textbf{69.16}} & 98.81 & \resulttwo{\textbf{0.27}} & \resulttwo{\textbf{63.41}} & \resultthird{\underline{55.68}} & \resulttwo{\textbf{66.08}} & \underline{56.58} \\
Qwen3-VL-30B-A3B~\cite{bai2025qwen3vltechnicalreport} \video & \underline{34.19} & \underline{52.09} & 38.54 & \resultthird{\textbf{14.46}} & \underline{57.26} & 73.91 & 1.18 & \resultthird{\underline{54.68}} & \resulttwo{\textbf{57.90}} & \resultthird{\underline{65.98}} & \resulttwo{\textbf{65.22}} \\ \hline
\multicolumn{12}{l}{\textit{Open-source Streaming Models}} \\ \hline
Dispider~\cite{qian2025dispider} \video & 15.44 & 21.71 & 19.29 & 8.09 & 23.90 & \resultthird{55.63} & 7.26 & 38.01 & 21.97 & 23.65 & 31.37 \\
VideoLLM-online-8B~\cite{chen2024videollm} \video & 8.48 & 15.00 & 15.53 & 0.03 & 17.67 & 99.10 & \resultthird{\underline{0.66}} & 13.15 & 16.90 & 10.15 & 6.70 \\
MMDuet2~\cite{wang2025mmduet2} \video & 6.79 & 11.76 & 10.37 & 1.44 & 11.27 & \resultone{\textbf{31.49}} & 54.11 & 15.84 & 14.44 & 9.16 & 4.96 \\ 
\bottomrule
\end{tabular}%
                       }
\vspace{-0.5cm}
\end{table}

\begin{table}[bhtp]
\centering
\caption{Comprehensive performance comparison of MLLMs across 3 multi-stream dimensions and 11 core tasks on $X$-Stream Benchmark.
The “$*$” symbol indicates that the model lacks audio capabilities and is evaluated directly on the question, while other notations follow Tab.~\ref{tab:main_comparison}.
}
\label{tab:main_comparison_core_task}
\vspace{-0.2cm}
\resizebox{\textwidth}{!}{%
\begin{tabular}{l|ccccc|ccccc|c}
\toprule
\rowcolor{gray!30}  & \multicolumn{5}{c|}{\textbf{Foundational Grounding}} & \multicolumn{5}{c|}{\textbf{Logical Cognition}} & \textbf{Agency} \\ \cline{2-12} 
\rowcolor{gray!30} \multirow{-2}{*}{\textbf{Model}}& \textbf{\begin{tabular}[c]{@{}c@{}}Visual\\ Grd.\end{tabular}} & \textbf{\begin{tabular}[c]{@{}c@{}}Audio\\ Grd.\end{tabular}} & \textbf{\begin{tabular}[c]{@{}c@{}}Temporal\\ Grd.\end{tabular}} & \textbf{\begin{tabular}[c]{@{}c@{}}Object\\ Count.\end{tabular}} & \textbf{\begin{tabular}[c]{@{}c@{}}Saliency\\ Detect.\end{tabular}} & \textbf{\begin{tabular}[c]{@{}c@{}}3D\\ spa.\end{tabular}} &  \textbf{\begin{tabular}[c]{@{}c@{}}Counter-\\factual \end{tabular}} & \textbf{\begin{tabular}[c]{@{}c@{}}Causal\\ Reasoning\end{tabular}} & \textbf{\begin{tabular}[c]{@{}c@{}}Common\\ Sense\end{tabular}} & \textbf{\begin{tabular}[c]{@{}c@{}}Anomaly\\ Detect.\end{tabular}} & \textbf{\begin{tabular}[c]{@{}c@{}}Decision\\ -Making\end{tabular}} \\ \hline
\multicolumn{12}{l}{\textit{Proprietary Multimodal Models}} \\ \hline
Gemini 3 Pro~\cite{gemini3pro2025} \video \space \audio & \resultone{66.72} & \resultone{64.82} & \resultone{68.93} & \resultone{76.37} & \resulttwo{63.61} & \resultone{70.82} & \resultthird{75.00} & \resultthird{41.79} & \resultone{69.35} & \resultone{70.52} & \resultone{44.18} \\
GPT-5~\cite{singh2025openai} \video & 36.68 & 21.63$^*$ & 36.64 & 42.52 & 53.55 & 52.28 & 15.00 & 37.65 & 44.77 & 45.54 & 28.74 \\
GPT-4o~\cite{hurst2024gpt} \video & 31.99 & 22.15$^*$ & 32.83 & 36.19 & 42.14 & 40.74 & 40.00 & 33.33 & 38.04 & 37.86 & 24.53 \\
Doubao-Seed-1.8~\cite{seed2026vision} \video & 49.87 & \resultthird{29.91} & 49.31 & 52.75 & \resultthird{61.13} & 59.14 & \resulttwo{85.00} & \resulttwo{52.45} & \resultthird{57.92} & \resultthird{54.29} & \resulttwo{37.10} \\ \hline
\multicolumn{12}{l}{\textit{Open-source Multimodal Models}} \\ \hline
Qwen2.5-VL-7B \video & 38.90 & 27.82$^*$ & 40.12 & 30.16 & 44.26 & 45.22 & 40.00 & 36.27 & 40.02 & 33.04 & 21.19 \\
Qwen3-VL-8B~\cite{bai2025qwen3vltechnicalreport} \video& 46.03 & 25.84$^*$ & 47.26 & 46.60 & 49.82 & 48.95 & 20.00 & 38.24 & 42.65 & 35.89 & \underline{30.18} \\
Qwen3-Omni-30B-A3B~\cite{xu2025qwen3omnitechnicalreport} \video \space \audio & \resultthird{\underline{53.61}} & \resulttwo{\textbf{52.15}} & \resulttwo{\textbf{64.56}} & \resulttwo{\textbf{64.77}} & \resultone{\textbf{68.61}} & \resulttwo{\textbf{63.29}} & \resultone{\textbf{90.00}} & \resultone{\textbf{53.14}} & \resulttwo{\textbf{64.08}} & \resulttwo{\textbf{60.18}} & 27.14 \\
Qwen3-VL-30B-A3B~\cite{bai2025qwen3vltechnicalreport} \video& \resulttwo{\textbf{64.33}} & \underline{29.83}$^*$ & \resultthird{\underline{54.55}} & \resultthird{\underline{54.68}} & \underline{60.96} & \resultthird{\underline{60.45}} & 40.00 & \underline{40.22} & \underline{50.84} & \underline{41.25} & \resultthird{\textbf{31.59}} \\ \hline
\multicolumn{12}{l}{\textit{Open-source Streaming Models}} \\ \hline
Dispider~\cite{qian2025dispider} \video & 19.58 & 13.80$^*$ & 18.67 & 25.10 & 22.50 & 22.12 & \underline{50.00} & 17.20 & 21.65 & 17.32 & 10.94 \\
VideoLLM-online-8B~\cite{chen2024videollm} \video & 12.62 & 17.97$^*$ & 21.47 & 12.90 & 22.64 & 21.94 & 0.00 & 18.14 & 21.53 & 21.13 & 14.44 \\
MMDuet2~\cite{wang2025mmduet2} \video & 22.27 & 18.35$^*$ & 22.24 & 36.10 & 24.14 & 23.63 & 10.00 & 17.90 & 22.16 & 20.42 & 9.12 \\ 
\bottomrule
\end{tabular}%
}
\vspace{-0.3cm}
\end{table}

\vspace{-0.2cm}
\subsection{Main Results}
\vspace{-0.1cm}
\label{sec:main_results}

\textbf{Performance comparison of streaming abilities.} 
In Tab.\ref{tab:main_comparison}, we adopt the straightforward Spatial Division Multiplexing for our primary comparative experiments. As shown in Tab.\ref{tab:main_comparison}, proprietary models consistently outperform their open-source counterparts across all settings on the full $X$-Stream benchmark. Notably, while Qwen3-Omni-30B-A3B leads the open-source models due to its strong comprehension, its overall Forward capability is hindered by suboptimal response timing. Furthermore, open-source streaming models generally underperform, constrained by limited training data and difficulties in handling frequent proactive queries. Consequently, we select the leading proprietary model, Gemini 3 Pro, alongside top-performing open-source models for the in-depth observational experiments detailed below.

\textbf{Performance comparison of multi-stream abilities.} 
In Fig.~\ref{tab:main_comparison_core_task}, $X$-Stream's three-dimensional evaluation reveals a clear multi-stream capability hierarchy across current MLLMs, transitioning from basic perception to complex reasoning. When calculating dimension subtask and ability scores, timing effects are excluded to prevent outliers from affecting the results, therefore, yielding a higher overall score. Models generally exhibit robust performance in foundation tasks, but struggle significantly with advanced cognitive demands. Specifically, decision-making and specific logical cognition tasks like causal reasoning emerge as the most formidable bottlenecks, yielding lower scores across all model tiers. Overall, $X$-Stream perfectly serves as the multi-stream evaluation standard.

\begin{table}[t]
    \centering
    
    \begin{minipage}[t]{0.5\textwidth}
        \centering
        \caption{The impact of multiplexing ablation on individual abilities on all $X$-Stream Benchmark.}
        \vspace{-0.2cm}
        \label{tab:comparison_individual_abilities}
        \begin{adjustbox}{width=\textwidth}
            \begin{tabular}{lc|cccc}
            \toprule
            \rowcolor{gray!30} \textbf{Model}                                               & \textbf{Scheme} & \textbf{\begin{tabular}[c]{@{}c@{}}Single\\ Stream\end{tabular}} & \textbf{\begin{tabular}[c]{@{}c@{}}Multi-\\ Coop\end{tabular}} & \textbf{\begin{tabular}[c]{@{}c@{}}Cross-\\ Ref\end{tabular}} & \textbf{\begin{tabular}[c]{@{}c@{}}Cross-\\ Inter\end{tabular}} \\ \hline
            \multirow{3}{*}{\begin{tabular}[c]{@{}l@{}}Qwen3-Omni\\ -30B-A3B\end{tabular}} & Spatial           & 63.41                                                           & 55.68                                                               & \textbf{66.08}                                                              & 56.58                                                                \\
                                                                                            & Time            & \textbf{69.78}                                                           & \textbf{58.76}                                                               & 58.62                                                              & \textbf{69.14}                                                                \\
                                                                                            & Semantic            & 61.58                                                           & 52.13                                                               & 55.34                                                              & 59.03                                                                \\ \hline
            \multirow{3}{*}{\begin{tabular}[c]{@{}l@{}}Gemini-3-\\ -Pro\end{tabular}}       & Spatial           & 72.45                                                           & 71.16                                                               & \textbf{74.79}                                                              & 66.96                                                                \\
                                                                                            & Time            & \textbf{79.62}                                                           & \textbf{75.13}                                                               & 67.08                                                              & \textbf{80.74}                                                                \\
                                                                                            & Semantic            & 70.30                                                           & 66.64                                                               & 70.92                                                              & 68.94           \\
            \bottomrule
            \end{tabular}
        \end{adjustbox}
    \end{minipage}\hfill
    \begin{minipage}[t]{0.48\textwidth}
        \centering
        \caption{Performance comparison of the number of streams under different multiplexing schemes.}
        \vspace{-0.2cm}
        \label{tab:comparison_number_streams}
        \begin{adjustbox}{width=\textwidth}
            \begin{tabular}{lc|cccc}
            \toprule
            \rowcolor{gray!30} \textbf{Model} & \textbf{Scheme} & \textit{N}=2 & \textit{N}=3 & \textit{N}=4 & \textit{N}=5 \\ \hline
            \multirow{3}{*}{\begin{tabular}[c]{@{}c@{}}Qwen3-Omni-\\ -30B-A3B\end{tabular}} 
                                              & Spatial                 & 36.61 & 36.88 & 34.22 & 25.84 \\
                                              & Time                  & \textbf{40.55} & 36.17 & 35.83 & 25.44 \\
                                              & Semantic                  & 36.15 & \textbf{38.46} & \textbf{40.64} & \textbf{29.82} \\ \hline
            \multirow{3}{*}{\begin{tabular}[c]{@{}c@{}}Gemini-3-\\ -Pro\end{tabular}}       
                                              & Spatial                 & 57.47 & 19.86 & 30.48 & 22.81 \\
                                              & Time                  & \textbf{58.09} & 24.19 & 31.76 & 22.14 \\
                                              & Semantic                  & 55.50 & \textbf{26.89} & \textbf{41.25} & \textbf{31.06} \\
            \bottomrule
            \end{tabular}
        \end{adjustbox}
    \end{minipage}
        
    \vspace{-0.5cm}
\end{table}

\Preliminary{2}{
In signal theory, each scheme presents inherent limitations. 1) Frequency division $\rightarrow$ waste of resources; 2) time division $\rightarrow$ latency and jitter; and 3) code division $\rightarrow$ interference between codes.
}
\vspace{0.3cm}

\textbf{Analysis of multiplexing scheme.} In a hypothetical scenario with a large number of streams, spatial or time division would degrade into full blurriness or discontinuity, losing all usable information; semantic division, however, would still manage to retain a basic semantic content. Conversely, in a single-stream scenario, the outcome of spatial and time division becomes standard streaming video, whereas semantic division would introduce unnecessary semantic loss. Based on our extended evaluations, we summarize the distinct advantages of each multiplexing approach:

\noindent\textbf{1)} \textbf{Spatial Division Multiplexing} excels in temporal modeling and cross-stream referencing. As shown in Tab.\ref{tab:comparison_individual_abilities}, it demonstrates superior cross-stream referencing in multi-stream setups. This likely occurs because representing multiple frames within a single unit preserves the model's pretrained inference dynamics and temporal perception.

\noindent\textbf{2)} \textbf{Time Division Multiplexing }is optimal under relaxed token rate constraints. As shown in Tab. \ref{tab:comparison_individual_abilities} and \ref{tab:comparison_single_stream}, evaluations reveal that this method thrives in dual-stream scenarios, which form the main part of our $X$-Stream Bench. Furthermore, by circumventing the severe selection penalties of Semantic Division, it better preserves visual details and delivers stronger performance in 2-stream scenarios. 

\noindent\textbf{3)} \textbf{Semantic Division Multiplexing} dominates under strict token constraints and high stream counts ($\ge 3$ streams). As Tab.\ref{tab:main_comparison_core_task} illustrates, scaling up the number of streams causes Spatial and Time Division representations to become severely blurred or fragmented. Under these dense information loads, Semantic Division effectively filters and preserves critical semantic features.

\noindent These empirical findings are \textbf{similar to} the time-tested theory of signal multiplexing. Our exploration of multi-stream video multiplexing can serve as the foundation of future multi-streaming understanding models.

\begin{table}[t]
    \centering
    
    \begin{minipage}[t]{0.35\textwidth}
        \centering
        \caption{Performance comparison of audio settings. In spatial division, we simply superimpose the audio tracks.}
        \label{tab:comparison_modalities}
        \begin{adjustbox}{width=\textwidth}
            \begin{tabular}{l|cccc} 
            \toprule
            \rowcolor{gray!30} \multirow{2}{*}{\textbf{Modality}} & \multicolumn{2}{c}{\textbf{\begin{tabular}[c]{@{}c@{}}Qwen3-Omni\\ -30B-A3B\end{tabular}}} & \multicolumn{2}{c}{\textbf{\begin{tabular}[c]{@{}c@{}}Gemini-3\\ -Pro\end{tabular}}} \\
            \rowcolor{gray!30}                           & Spatial                                        & Time                                         & Spatial               & Time                \\ \hline
            {\textit{w/.} audio}                  & \textbf{34.28}                                        & 26.57                                        & \textbf{49.60}               & 41.52               \\
            {\textit{w/o.} audio}                 & 29.40                                        & \textbf{30.37}                                        & 46.13               & \textbf{45.84}               \\ \bottomrule
            \end{tabular}
        \end{adjustbox}
    \end{minipage}\hfill
    \begin{minipage}[t]{0.62\textwidth}
        \centering
        \caption{Performance comparison in multi-stream setting. Videos from single-stream datasets are combined with distractor videos to form  pseudo multi-stream.
        }
        \label{tab:comparison_single_stream}
        \begin{adjustbox}{width=\textwidth}
            \begin{tabular}{lc|c|ccc}
            \toprule
             \rowcolor{gray!30}   &  & \textbf{Multi-Stream}                                & \multicolumn{3}{c}{\textbf{Single Stream}}                                                                                                                                  \\
\rowcolor{gray!30}  \multirow{-2}{*}{\textbf{Model}} 
& \multirow{-2}{*}{\textbf{Input}}    & \begin{tabular}[c]{@{}c@{}}X-Stream\\(Ours) \end{tabular} & \begin{tabular}[c]{@{}c@{}}Streaming-\\ -Bench\end{tabular} & \begin{tabular}[c]{@{}c@{}}OVO\\ -Bench\end{tabular} & \begin{tabular}[c]{@{}c@{}}Pho-\\ -Stream\end{tabular} \\ \hline
            \multirow{2}{*}{\begin{tabular}[c]{@{}l@{}}Qwen3-Omni\\ -30B-A3B\end{tabular}} & Single Stream                   & 5.35                                                 & \textbf{57.19}                                                       & \textbf{56.16}                                                & \textbf{33.01}                                                  \\
                                                                                           & Multiple Stream                 & \textbf{34.28}                                                & 26.10                                                       & 34.78                                                & 17.33                                                  \\ \hline
            \multirow{2}{*}{Gemini 3 Pro}                                                  & Single Stream                   & 13.93                                                 & \textbf{65.20}                                                       & \textbf{58.91}                                                & \textbf{44.64}                                                  \\
                                                                                           & Multiple Stream                 & \textbf{49.60}                                                & 35.91                                                       & 38.05                                                & 13.96                                                  \\
            \bottomrule
            \end{tabular}
        \end{adjustbox}
    \end{minipage}
    \vspace{-0.5cm}
\end{table}

\textbf{Analysis of multiplexing ablation on audio.} 
Tab.~\ref{tab:comparison_individual_abilities} shows that audio-capable methods perform significantly better in audio grounding, highlighting the importance of audio. However, Tab.\ref{tab:comparison_modalities} reveals that the impact of multiplexing varies significantly between audio and video. For instance, Spatial Division Multiplexing often causes multi-channel audio to overlap. Conversely, while Time-Division resolves this overlap, it introduces semantic discontinuity in speech. Since audio signals possess stronger inherent coupling than image pixels, our discussion is limited to simple multiplexing techniques.

\textbf{Necessity of multi-stream.}
Tab.\ref{tab:comparison_single_stream} presents further results validating the necessity of multi-stream processing. Single-stream inference fails on our multi-stream benchmark. Conversely, injecting distracting streams into single-stream datasets causes severe performance degradation. This confirms that our $X$-Stream benchmark evaluates a fundamentally novel capability, rather than merely extending single-stream systems.

\textbf{Human verification of LLM-as-judge}:
To validate the effectiveness, we request both the LLM-as-judge and human experts to evaluate 200 QAs. The Spearman correlation of 0.62 ($p$<0.05) confirms that LLM-as-judge reliably mirrors human evaluation. See the appendix for prompt and model details.

\begin{figure}[t]
\vspace{-0.1cm}
\centering
    \includegraphics[width=0.95\textwidth]{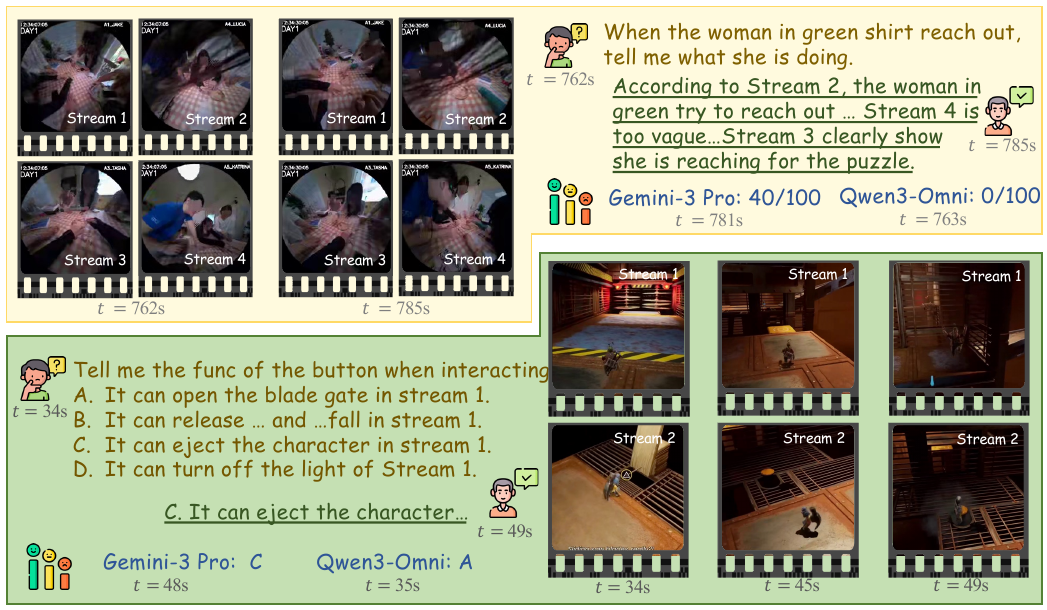}
\vspace{-0.2cm}
\caption{The case study in our $X$-Stream Benchmark. We choose a 4-stream, proactive, free-form QA (yellow) and a 2-stream, proactive, multi-choice QA (green) as examples.}
\vspace{-0.5cm}
\label{fig:two_cases}
\end{figure}

\textbf{More experiments and analysis.} As shown in Fig.\ref{fig:two_cases}, our $X$-Stream benchmark requires the integration of multi-stream information to generate timely and accurate answers.
We encourage readers to consult the appendix for more details, including case preview, experiments, and analysis.

\vspace{-0.2cm}
\section{Discussion and Conclusion}
\label{sec:conclusion}
\vspace{-0.2cm}

\textbf{Discussion}: 
Despite its wide range of applications, the full potential of multi-stream remains untapped.
On the data front, public video datasets are not precise enough without film-standard professional gear and expert work, usually causing streams to drift out of sync over time. On the technical side, current multiplexing strategies still struggle to balance video comprehension with temporal reasoning.

\noindent\textbf{Conclusion}: 
In this paper, we introduce X-Stream, the first comprehensive benchmark dedicated to multi-stream streaming understanding. To cover as many real-world scenarios as possible, we develop a rigorous dual-verification pipeline, ensuring that our 4,220 curated QA pairs genuinely demand cross-stream understanding. By evaluating popular MLLMs as naive multiplexers, our extensive experiments reveal that current models still struggle with continuous multi-stream integration, achieving only around 50\% accuracy and falling short in proactive tasks. Furthermore, we systematically analyze the inherent trade-offs of three multiplexing strategies under varying token constraints and stream counts. Ultimately, X-Stream exposes the limitations of existing streaming architectures, providing both a robust evaluation framework and empirical guidance for designing the next generation of efficient, real-time multi-stream agents.

\section*{Acknowledgments}

This work is partially supported by HK RGC-Early Career Scheme (No. 24211525), and ITSP Platform Project (No. ITS/600/24FP). This study was supported in part by the Centre for Perceptual and Interactive Intelligence, a CUHK-led InnoCentre under the InnoHK initiative of the Innovation and Technology Commission of the Hong Kong Special Administrative Region Government. This work is also partially supported by Hong Kong RGC Strategic Topics Grant (No. STG1/E-403/24-N), and a CUHK-CUHK(SZ)-GDST Joint Collaboration Fund (No. YSP26-4760949). This work is also partially supported by the Huawei Hong Kong Research Center (HKRC).

\bibliographystyle{splncs04}
\bibliography{main}

\newpage
\appendix
\renewcommand\thefigure{\Alph{section}\arabic{figure}}
\renewcommand\thetable{\Alph{section}\arabic{table}} 

\vspace{0.5cm}

In this supplementary material, we provide two key components. 

\begin{itemize}[label=\textbf{$\bullet$}]
    \item we release a preview version of the \textbf{evaluation code} in the attached compressed file. 
    \item we include \textbf{additional information} for the reader's reference, such as data sources, data analysis, data previews, and empirical observations.
\end{itemize}

As the \textbf{first benchmark for multi-stream understanding}, $X$-Stream establishes a comprehensive and effective evaluation framework for measuring the ability of existing models to perceive, understand, and reason across multiple streams.

\section*{Appendix Contents}

\startcontents[appendices]
\setcounter{tocdepth}{2}
\printcontents[appendices]{}{2}{}

\newpage

\section{Multi-Stream Data Preview}

\vspace{-0.2cm}

\begin{figure}[htbp]
\centering
    \includegraphics[width=0.95\textwidth]{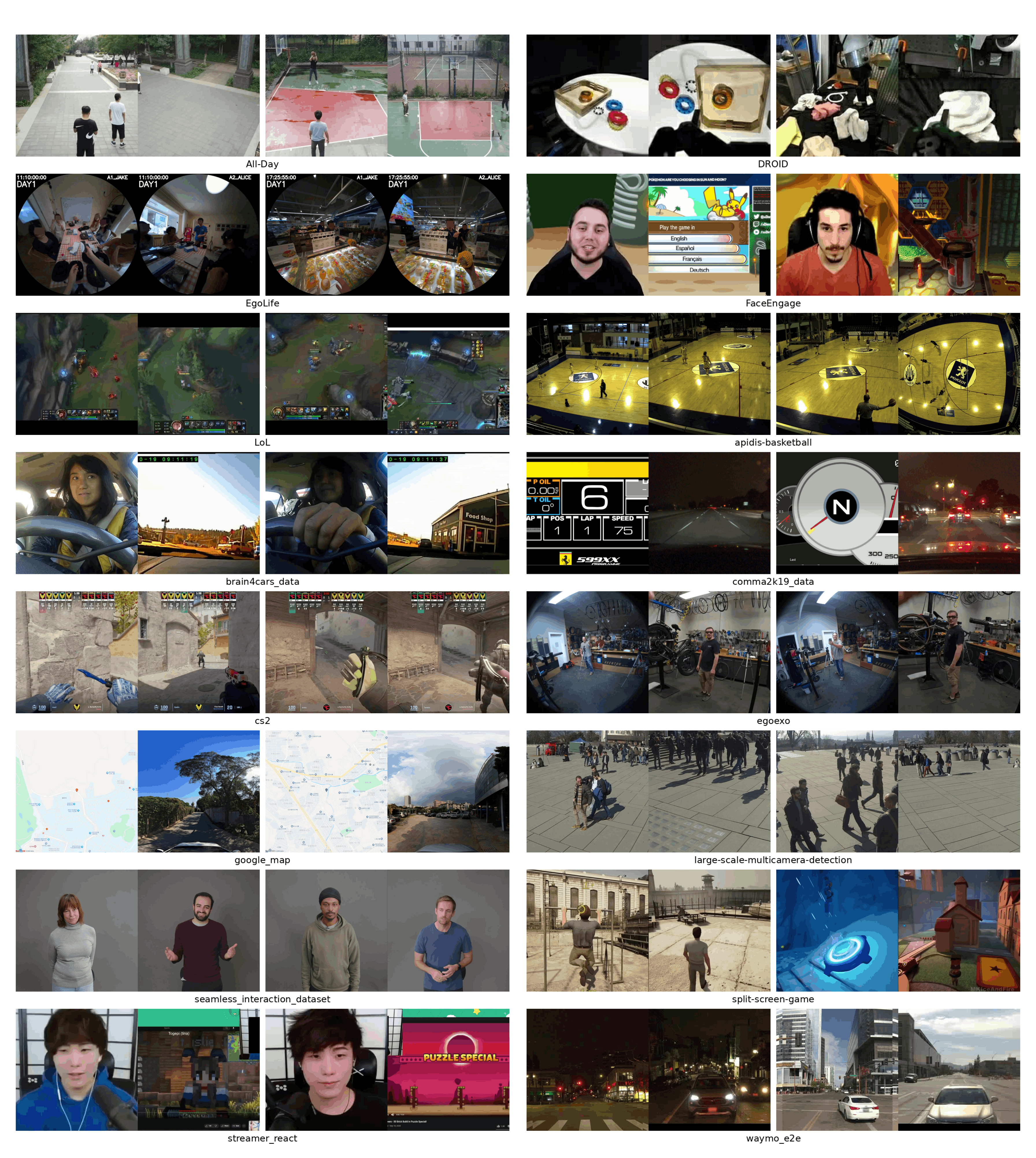}
\vspace{-0.5cm}
\caption{Data Preview. This preview highlights the main real-world multi-stream applications and offers an overview of the diversity of our $X$-Stream.}
\label{fig:datapreview}
\end{figure}

Beyond the challenge of single continuous streams, modern real-world perception increasingly demands coordinated multi-stream collaboration across heterogeneous devices. Such collaboration underpins a remarkably wide range of applications, including multi-screen coordination in office environments, orchestration of live feeds in sports broadcasting, cooperative navigation between mobile maps and smart glasses, and the synchronization of shoulder and wrist cameras on robotic arms.
As the first benchmark dedicated to multi-stream streaming, our $X$-Stream encompasses a diverse set of real-world scenarios with rich multi-angle, multi-view, and multi-device characteristics.
Fig.~\ref{fig:datapreview} presents a preview of the multi-stream data in our main dataset.

\vspace{-0.5cm}

\section{Details of the $X$-Stream Benchmark}

\subsection{Data Sources}
As shown in Tab.~\ref{tab:data_sources_statistics}, we collected approximately 857 hours of data from 20 methods and sources. After a rigorous screening process, we ultimately retained about 160 hours of final data to construct our benchmark, in Tab.\ref{tab:eval_data_statistics}.

\begin{table}[!htbp]
  \centering
  \vspace{-5mm}

  \begin{minipage}{\textwidth}
    \centering
    \captionof{table}{Statistics of multi-domain video sources used in our study.}
    \label{tab:data_sources_statistics}
    \resizebox{0.75\textwidth}{!}{%
      \setlength\tabcolsep{6pt}
      \begin{tabular}{lcccc}
        \toprule
        \rowcolor{gray!30}
        \multirow{1}{*}{\textbf{Data Source}} & \multicolumn{1}{c}{\textbf{Takes Num.}} & \multicolumn{1}{c}{\textbf{Streams Num.}} & \multicolumn{1}{c}{\textbf{Hours}} & \multicolumn{1}{c}{\textbf{FPS}} \\
        \midrule
        \rowcolor{gray!30} \multicolumn{5}{l}{\textit{Driving}} \\
        brain4cars~\cite{jain2016brain4cars} & 594 & 2 & 2 & 30 \\
        Waymo-E2E~\cite{xu2025wod} & 1498 & 2 & 12 & 30 \\
        \midrule
        \rowcolor{gray!30} \multicolumn{5}{l}{\textit{Sports}} \\
        Apidis-Basketball~\cite{VanZandycke_DeepSport} & 164 & 7 & 19 & 30 \\
        e-Sports (Self-record) & 10 & 2--10 & 26 & 30 \\
        Split-screen Game (Youtube) & 35 & 2 & 9 & 30 \\
        \midrule
        \rowcolor{gray!30} \multicolumn{5}{l}{\textit{Robot}} \\
        DROID~\cite{khazatsky2024droid} & 26{,}000 & 2 & 138 & 30 \\
        UAV-loc-dataset~\cite{xu2024uav} & 11 & 2 & 2 & 1 \\
        \midrule
        \rowcolor{gray!30} \multicolumn{5}{l}{\textit{Daily Routine}} \\
        EgoExo4D~\cite{grauman2024ego} & 3{,}800 & 4--6 & 250 & 30 \\
        EgoLife~\cite{yang2025egolife} & 42 & 6 & 138 & 30 \\
        \midrule
        \rowcolor{gray!30} \multicolumn{5}{l}{\textit{Chat}} \\
        Seamless-interaction~\cite{agrawal2025seamless} & 1,322 & 2 & 143 & 30 \\
        \midrule
        \rowcolor{gray!30} \multicolumn{5}{l}{\textit{Surveillance}} \\
        WILDTRACK~\cite{chavdarovawildtrack} & 1 & 7 & 4 & 60 \\
        All-Day~\cite{fan2025all} & 19 & 2 & 2 & 30 \\
        \midrule
        \rowcolor{gray!30} \multicolumn{5}{l}{\textit{Live Streaming}} \\
        FaceEngage~\cite{chen2019faceengage} & 25 & 2 & 2 & 30 \\
        Streamer-React (Youtube) & 26 & 1 & 6 & 24 \\
        \midrule
        \rowcolor{gray!30} \multicolumn{5}{l}{\textit{Interfaces}} \\
        Map-Street (Baidu/Google Map API) & 213 & 2 & 61 & 1 \\
        Comma2K-19~\cite{schafer2018commute} w/. dashboard & 2,037 & 2 & 63 & 10 \\
        \midrule
        \rowcolor{gray!30} \multicolumn{5}{l}{\textit{Total}} \\
        Multi-Stream & - & - & 857 & - \\
        \bottomrule
      \end{tabular}
    }
  \end{minipage}

  \vspace{4mm}

  \begin{minipage}{\textwidth}
    \centering
    \captionof{table}{Statistics of video source in $X$-Stream.}
    \label{tab:eval_data_statistics}
    \resizebox{0.75\textwidth}{!}{%
      \setlength\tabcolsep{6pt}
      \begin{tabular}{lccc}
        \toprule
        \rowcolor{gray!30}
        \textbf{Data Source} & \textbf{Takes Count} & \textbf{Shot Hours} & \textbf{Video Hours} \\
        \midrule
        All-Day & 13 & 0.77 & 1.74 \\
        DROID & 62 & 11.42 & 22.95 \\
        UAV-loc-dataset & 3 & 0.73 & 1.46 \\
        EgoLife & 50 & 9.43 & 28.27 \\
        FaceEngage & 40 & 1.49 & 2.99 \\
        LoL & 13 & 2.28 & 6.25 \\
        apidis-basketball & 15 & 1.70 & 3.39 \\
        brain4cars & 12 & 1.09 & 2.17 \\
        comma2k19 & 60 & 11.14 & 22.17 \\
        cs2 & 35 & 5.79 & 17.41 \\
        egoexo & 19 & 2.69 & 5.38 \\
        google-map & 29 & 5.68 & 11.31 \\
        large-scale-multicamera-detection & 10 & 1.03 & 3.38 \\
        seamless-interaction-dataset & 33 & 6.05 & 12.10 \\
        split-screen-game & 29 & 5.29 & 10.58 \\
        streamer-react & 8 & 1.04 & 2.08 \\
        waymo-e2e & 23 & 3.23 & 6.46 \\
        \midrule
        \rowcolor{gray!30} Total & 451 & 70.90 & 160.30 \\
        \bottomrule
      \end{tabular}
    }
  \end{minipage}

\end{table}

\subsubsection{Live Streaming Data}
Reaction videos with face cam and screen created by live streamers, are a representative example of the core interaction patterns in multi-stream experiences, as they naturally capture real-time responses to shared video or gameplay content. To construct our data source, we collected reaction clips from YouTube featuring the top 10 streamers by popularity: \texttt{iShowSpeed}, \texttt{Kai Cenat}, \texttt{Tyler1}, \texttt{xQc}, \texttt{Pokimane}, \texttt{HasanAbi}, \texttt{Ludwig}, \texttt{Sykkuno}, \texttt{Valkyrae}, and \texttt{Asmongold}. We focused specifically on videos in which these creators react to either online videos or game-related content, and we constrained the clip duration to between 5 and 30 minutes to ensure consistency and comparability across samples.

\subsubsection{Multi-Stream Game Data}
For the multi-stream gameplay component, we curated a set of widely played titles to cover different interaction and viewing dynamics. We grouped games into two categories: competitive esports-style games (\texttt{Counter-Strike 2}, \texttt{Mario Kart 8}, and \texttt{League of Legends}) and other games (\texttt{A Way Out}, \texttt{It Takes Two}, and \texttt{Split Fiction}). For each selected title, we sourced gameplay-only videos from YouTube—i.e., recordings that primarily contain the game feed with minimal overlays and without streamer face-cams or reaction framing—to serve as standardized base streams. We then compiled these gameplay streams per title and category for downstream processing and analysis in the multi-stream setting.

In cases where multi-view streams were difficult to obtain directly (e.g., when multiple player perspectives were unavailable on public platforms), we record the required viewpoints ourselves using a controlled capture pipeline. Specifically, for \textit{Source Engine (Valve Inc.)} we leveraged \textit{Half-Life Advanced Effects (HLAE)} to programmatically control the playback, enabling systematic extraction of multiple gameplay perspectives from the same session. We then used OBS to record and rectify the resulting feeds into standardized video outputs, ensuring consistent framing and quality across all captured viewpoints.

\vspace{-0.2cm}
\subsubsection{Car view with Dashboard}
\vspace{-0.2cm}
We construct a paired video-telemetry dataset using driving segments from the comma2k19~\cite{schafer2018commute} dataset. To synchronize multimodal data, we evaluated several simulators and ultimately selected the professional simulator SimHub\footnote{https://www.simhubdash.com/}, which uses the OutGauge LFS protocol and UDP for data transmission during simulation. Therefore, we establish CAN speed timestamps as the reference time axis and interpolate all other sensor signals (e.g., steering angle, wheel speed) onto this axis. Video frames are then aligned to the telemetry sequence by matching each frame to the nearest temporal state sample. To ensure a comprehensive vehicle-state representation, missing control variables—such as throttle, brake, gear, and RPM—are analytically estimated from longitudinal speed and acceleration. The resulting time-aligned state vectors and corresponding video data are explicitly paired to facilitate downstream tasks.

\vspace{-0.2cm}
\subsubsection{Map and street view}
\vspace{-0.2cm}
To construct our paired street-view and map dataset, we first sample random origin and destination coordinates (BD09) with guaranteed Baidu Street View coverage. Using the Baidu Direction API, we generate driving routes between these endpoints and densify them into a sequence of equidistant waypoints with computed headings. For each waypoint, we retrieve a street-view panorama and a corresponding static map image. After filtering out duplicate or incomplete frames to ensure strict one-to-one alignment, we output the paired modalities as synchronized videos, accompanied by a JSON file containing per-frame GPS and routing metadata. We will look into any subsequent licensing issues separately through our own inquiry and research.

\vspace{-0.4cm}
\subsection{Taxonomy of Cross-Stream Reasoning Tasks}
\vspace{-0.2cm}
In Tab.~\ref{tab:taxonomy}, we present a taxonomy of cross-stream reasoning tasks and their core logic, characterizing how information from multiple streams is combined, contrasted, and linked to produce an answer. The taxonomy groups tasks into four categories: (i) \textbf{cross-stream interference}, which probes redundancy and informativeness via stream ablation or corruption; (ii) \textbf{multi-stream cooperation}, which leverages complementary cues and explicit cross-stream comparison; (iii) \textbf{cross-stream reference}, which focuses on localization and temporal causal linking of events across views; and (iv) \textbf{Single-stream Understanding}, which uses only one stream to answer the question. Together, these four categories capture the dominant ways heterogeneous streams interact, providing a systematic lens on multimodal understanding and decision-making.

\begin{table*}[htbp]
\centering
\vspace{-0.4cm}
\caption{Taxonomy of Cross-stream Tasks and Core Logic. We organize task types according to the four core capabilities defined in X-Stream, based on how information from different streams contributes to the final answer.}
\label{tab:taxonomy}
\vspace{-0.2cm}
{\tiny
\resizebox{\textwidth}{1.15\height}{
\begin{tabularx}{\textwidth}{>{\raggedright\arraybackslash}p{2.2cm} >{\raggedright\arraybackslash}p{2.7cm} >{\raggedright\arraybackslash}p{2.7cm} >{\raggedright\arraybackslash}X}
\toprule
\rowcolor{gray!30} \textbf{Category} & \textbf{Sub-category} & \textbf{Core Logic} & \textbf{Typical Examples (Scenario)} \\
\midrule

\multirow{2}{=}{\textbf{1. Cross-stream Interference}} 
& Noise Filtering 
& Target stream A + distracting stream B $\to$ answer from A 
& In a split-screen setting, identify what the blue-haired girl in Stream 2 needs to do while ignoring visually salient but irrelevant actions in Stream 1. \\ 
\cmidrule(l){2-4}
& Contradiction Suppression
& Relevant cue in A + misleading cue in B $\to$ robust answer
& Determine whether pressing the triangle button is necessary in Stream 1 while avoiding confusion from similar controller actions shown in another stream. \\
\midrule

\multirow{2}{=}{\textbf{2. Multi-stream Cooperation}} 
& Complementary Reasoning 
& Clue in A + clue in B $\to$ answer 
& Did the driver looking at the phone (\textit{Inner}) cause the lane deviation (\textit{Outer})? \\ 
\cmidrule(l){2-4} 
& Multi-stream Evidence Aggregation 
& Partial evidence from A and B $\to$ joint conclusion 
& Detect an abnormal event only after combining surveillance footage from two different viewpoints. \\ 
\midrule

\multirow{2}{=}{\textbf{3. Cross-stream Reference}} 
& Cross-view Localization 
& Object/entity in A $\to$ corresponding object/entity in B 
& Where is the screw seen in the robotic arm view located in the global view? \\ 
\cmidrule(l){2-4}
& Temporal / Event Alignment 
& Event in A $\leftrightarrow$ event or state in B 
& What facial expression (\textit{Player}) was caused by the character's death (\textit{Game})? \\ 
\midrule

\multirow{2}{=}{\textbf{4. Single-stream Understanding}} 
& Stream-specific Perception 
& Query specifies one stream within a multi-stream context $\to$ answer from that stream 
& In Stream 2, what is the woman holding when she enters the room? \\ 
\cmidrule(l){2-4}
& Local Grounding in Context 
& Grounding/ counting/ recognition in A while other streams are present 
& In Stream 1, how many buttons are visible on the control panel at the queried moment? \\ 
\bottomrule
\end{tabularx}
}
}
\vspace{-0.6cm}
\end{table*}

\vspace{-0.4cm}
\subsection{Scenario-Specific QA Tasks}
\vspace{-0.2cm}
To avoid generating trivial questions, we designed practical QA tasks tailored to real-world scenarios, ensuring each provides distinct value within its specific application domain. Tab. \ref{tab:task_specific} summarizes these multi-stream scenarios and demonstrates how cross-stream reasoning supports complex question answering. We group the scenarios into three broad categories: multi-angle observation of the same entity, multi-view understanding of the same behavior, and multi-device collaboration for the same goal. For each case, the table details the scenario type, the primary understanding task, and an illustrative QA pair that highlights how synthesizing information across multiple streams yields a more accurate and complete answer.

\begin{table}[hbtp]
    \centering
    \tiny
    \renewcommand{\arraystretch}{1.12}
    \setlength{\tabcolsep}{4pt}
    \vspace{-0.3cm}
    \caption{\textbf{Representative settings given to LLMs as few-shot learners.}}
    \vspace{-0.3cm}
    \label{tab:task_specific}
    \resizebox{\textwidth}{1.2\height}{
        \begin{tabularx}{\textwidth}{@{} p{0.21\linewidth} p{0.23\linewidth} X @{}}
        \toprule
        \rowcolor{gray!30} \textbf{Stream Setting} & \textbf{Primary Task} & \textbf{Illustrative Cross-Stream Question} \\
        \midrule
        
        \multicolumn{3}{l}{\textit{\textbf{1. Different angle of the same object}}} \\
        \midrule
        
        \textbf{Robotics} (shoulder + wrist view) &
        Manipulation failure diagnosis &
        Why did the robot fail to insert the plug? The shoulder view shows that the arm reached the socket area, while the wrist view reveals that the plug was slightly misaligned. \\
        
        \textbf{Egocentric video} (ego + exo view) &
        Referring expression resolution &
        Which ingredient is the chef pointing to? The external view captures the pointing gesture, and the egocentric view identifies the jar near the fingertip as cumin. \\
        
        \textbf{Sports analysis} (side + goal-line view) &
        Rule and event verification &
        Was the goal legal? The side view establishes the passing moment, and the goal-line view helps verify the receiver's position relative to the defender. \\
        
        \textbf{Surveillance} (camera 1 + camera 2) &
        Cross-camera re-identification &
        Where did the person carrying the red bag go? One camera identifies the target, and the second camera captures the same individual entering the north-wing elevator shortly afterward. \\
        
        \midrule
        
        \multicolumn{3}{l}{\textit{\textbf{2. Different views of the same behavior}}} \\
        \midrule
        
        \textbf{Autonomous driving} (front + rear view) &
        Causal explanation of driving decisions &
        Why did the vehicle avoid changing lanes despite an open front view? The forward camera shows a clear lane, whereas the rear camera reveals a fast-approaching ambulance in the blind spot. \\
        
        \textbf{Collaborative gaming} (player 1 + player 2 view) &
        Team situational awareness &
        Did player 2 notice the enemy who eliminated player 1? One stream shows the attack direction, while the other indicates that player 2 was looking elsewhere at the same moment. \\
        
        \textbf{Social interaction} (participant 1 + participant 2 view) &
        Reaction grounding &
        What triggered the woman's laughter? One view captures her reaction, while the other shows her partner holding up a humorous drawing. \\
        
        \textbf{In-cabin monitoring} (road + driver view) &
        Driver awareness assessment &
        Did the driver notice the pedestrian? The road-facing camera records the pedestrian, while the in-cabin view shows the driver looking down at a phone. \\
        
        \midrule
        
        \multicolumn{3}{l}{\textit{\textbf{3. Different devices of the same goal}}} \\
        \midrule
        
        \textbf{Geo-localization} (street view + map) &
        Visual entity linking &
        What is the name of the company located in the blue building? The street view identifies the building, and the map stream links its location to the corresponding business entry. \\
        
        \textbf{Aerial inspection} (drone + satellite view) &
        Structural condition assessment &
        Is the bridge safe after the flood? The satellite view provides the broader layout, while the drone view reveals local cracks that indicate potential damage. \\
        
        \textbf{Vehicle status understanding} (road view + dashboard) &
        Context-aware alert explanation &
        Is the vehicle exceeding the legal speed limit? The dashboard reports the current speed, and the road view captures the posted speed-limit sign. \\
        
        \textbf{Game streaming} (gameplay + player cam) &
        Attribution of skill vs. chance &
        Was the headshot a matter of luck or skill? The gameplay stream shows the outcome, and the player camera provides evidence of a deliberate and rapid mouse movement. \\
        
        \bottomrule
    \end{tabularx}
    }

\vspace{-0.3cm}
\end{table}

\vspace{-0.4cm}
\subsection{Dataset Statistics}
\begin{figure}[htp]
\vspace{-0.5cm}
\centering
    \includegraphics[width=0.65\textwidth]{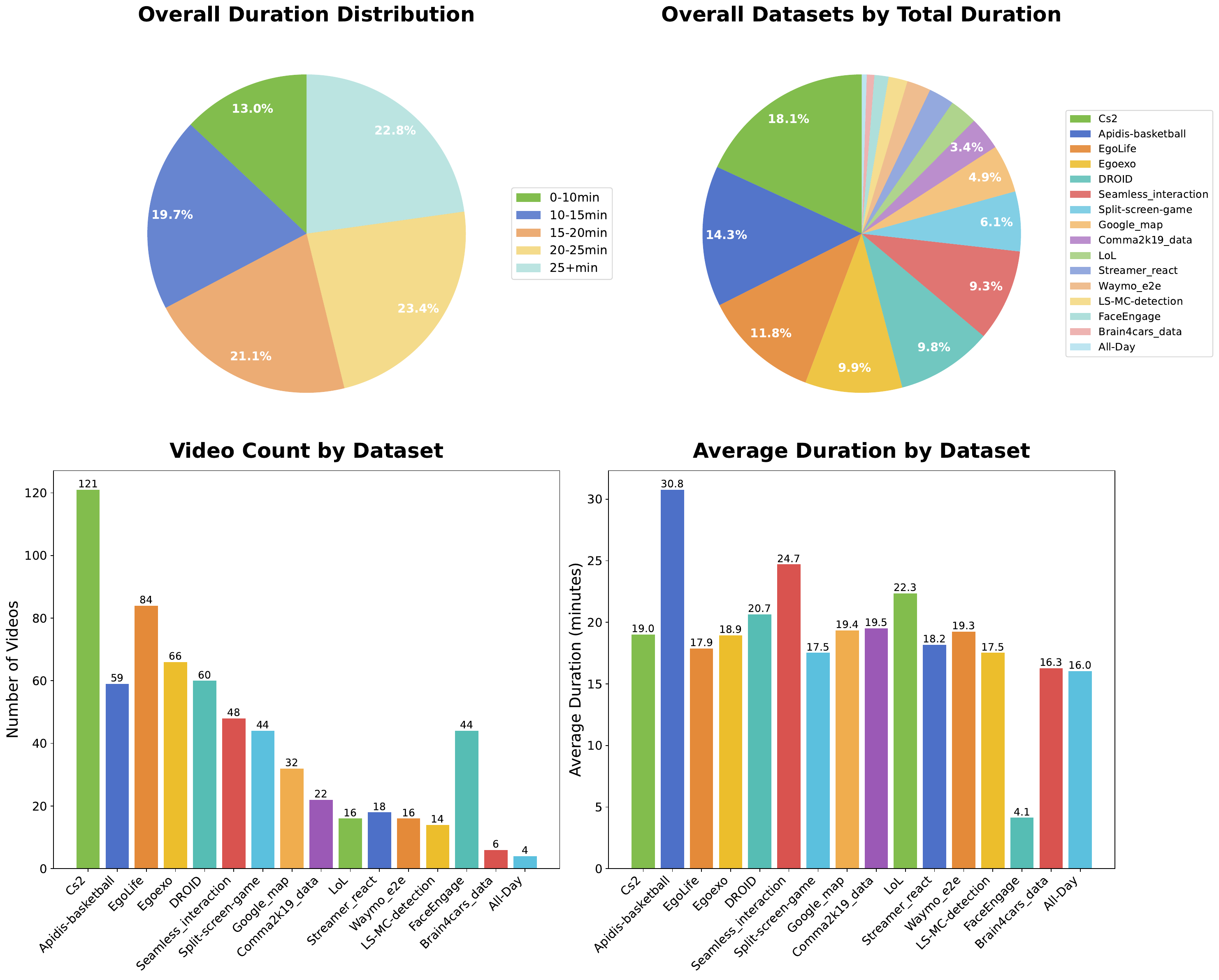}
\vspace{-0.3cm}
\caption{Distribution of the original data (before processing).}
\vspace{-0.5cm}
\label{fig:statistics_1}
\end{figure}

\begin{figure}[htp]
\vspace{-0.5cm}
\centering
    \includegraphics[width=0.6\textwidth]{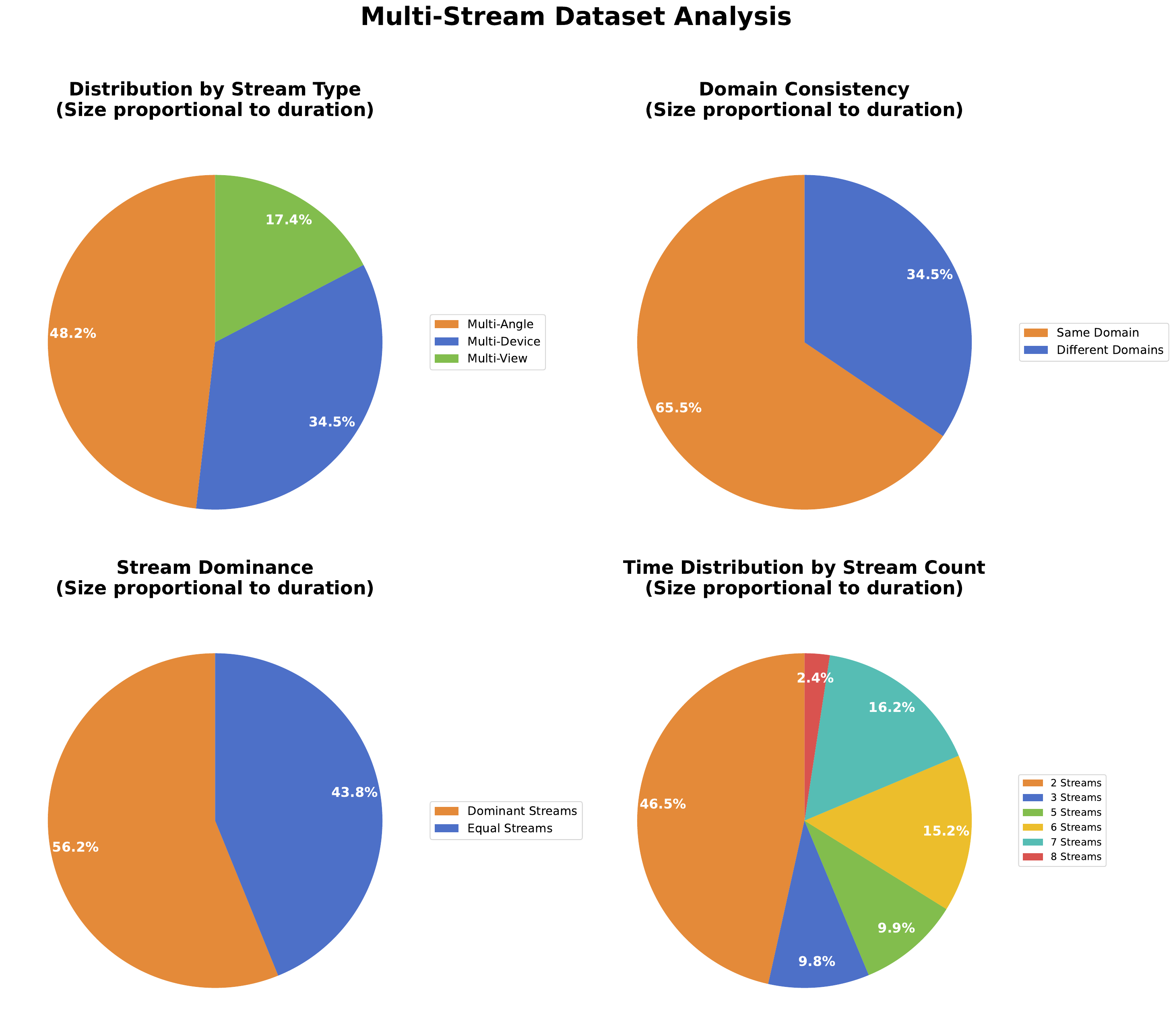}
\vspace{-0.5cm}
\caption{Distribution of the original data (before processing).}
\vspace{-0.5cm}
\label{fig:statistics_2}
\end{figure}

\vspace{-0.1cm}
We report key statistics of the original data (before processing) in the Fig.~\ref{fig:statistics_1} and Fig.~\ref{fig:statistics_2}.
The original data were collected with an emphasis on broader distribution coverage and greater diversity aligned with real-world scenarios.
Before processing, the original data were designed to cover a broad range of distributions and preserve the diversity of real-world scenarios as much as possible. As shown in the figure, the dataset exhibits substantial variation in duration, video count, stream type, domain consistency, and stream count. This broad coverage helps the data better reflect practical conditions and supports a more comprehensive evaluation.

\vspace{-0.5cm}

\begin{figure}[htp]
\centering
\vspace{-0.5cm}
\resizebox{0.8\textwidth}{!}{\includegraphics{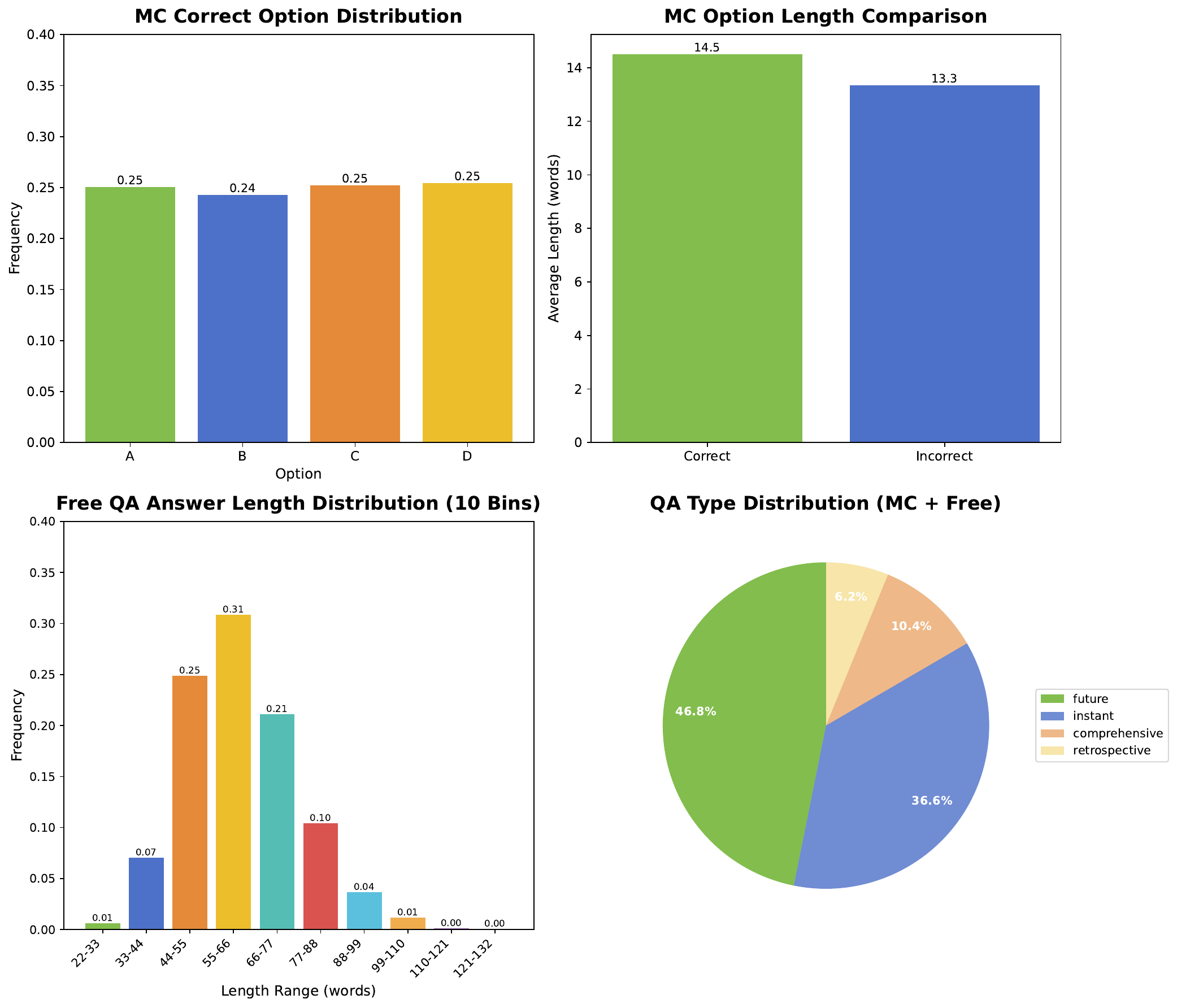}}
\vspace{-0.2cm}
\caption{Distribution of questions (after processing).}
\label{fig:question_distribution}
\end{figure}

\begin{figure}[htp]
\centering
\vspace{-0.6cm}
\resizebox{0.53\textwidth}{!}{\includegraphics{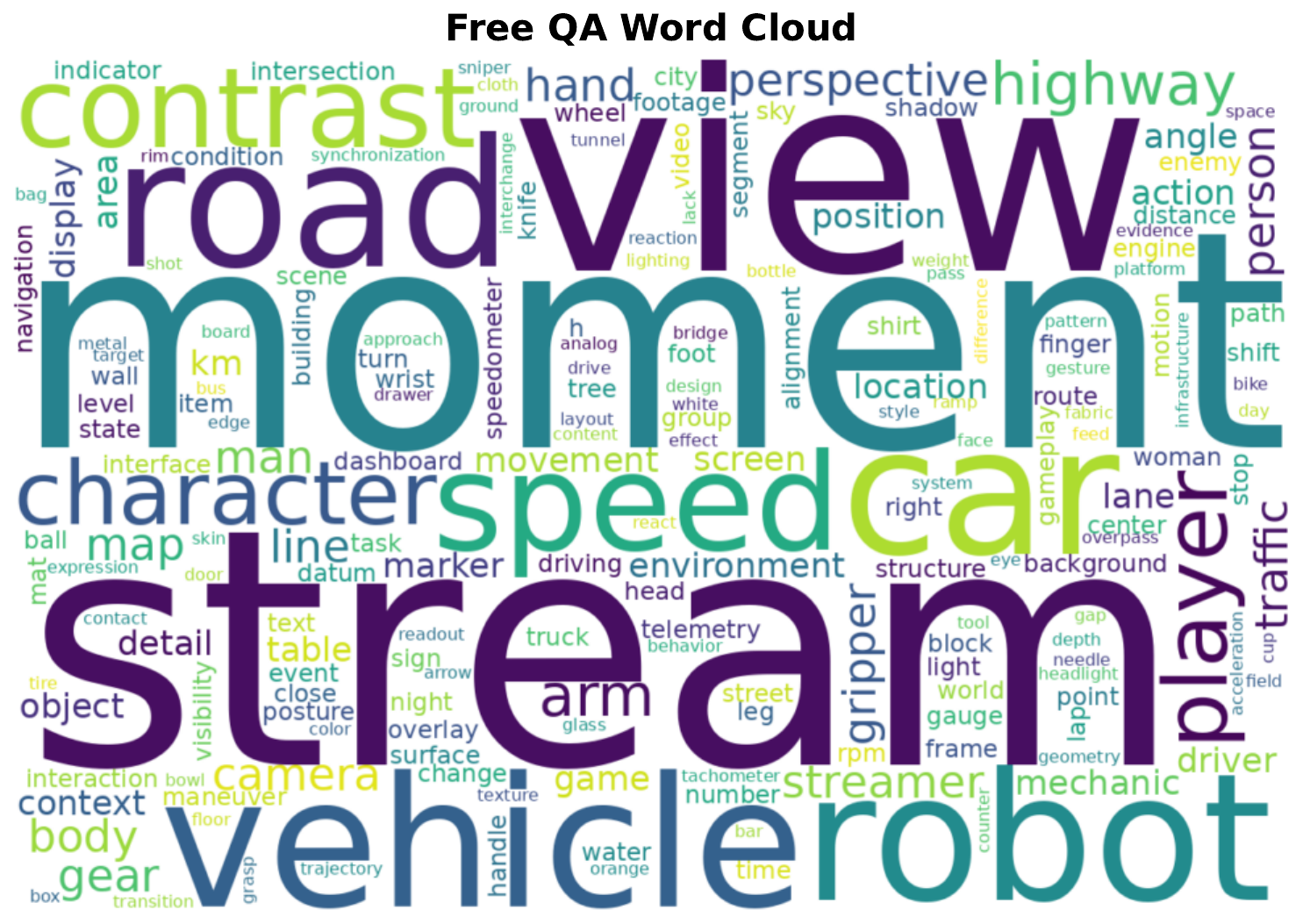}}
\vspace{-0.2cm}
\caption{Word cloud of the free-form answers (after processing).}
\vspace{-0.3cm}
\label{fig:freeform_wordcloud}
\end{figure}

Fig.~\ref{fig:question_distribution} and Fig.~\ref{fig:freeform_wordcloud} summarize the key statistics of the final dataset after processing, including the distributions of question types, answer lengths, and multiple-choice options. Overall, the dataset exhibits a relatively balanced composition across different categories, while also covering a diverse range of free-form answers. This distribution suggests that the dataset contains both structured multiple-choice questions and open-ended responses, which may help support a more comprehensive evaluation of model performance across different answering formats.

\vspace{-0.3cm}

\subsection{Details of Human Annotators}

In both the benchmark data annotation stage and the human test stage of our experiments, we hire 31 expert annotators with experience in multimodal video understanding. Annotators are compensated at a rate of \$18 per hour.

\begin{figure}[!htbp]
    \vspace{-0.6cm}
    \centering
    \begin{subfigure}[b]{0.48\textwidth}
        \centering
        \includegraphics[width=\textwidth]{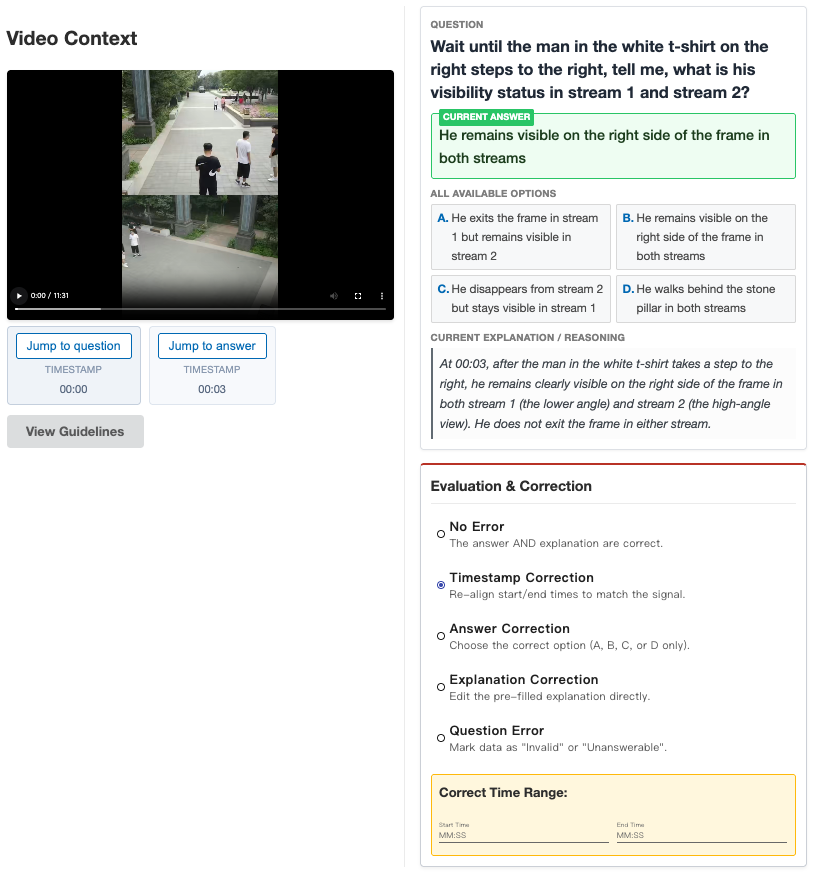}
        \caption{Timestamp correction case.}
        \label{fig:mturk1}
    \end{subfigure}
    \hfill
    \begin{subfigure}[b]{0.48\textwidth}
        \centering
        \includegraphics[width=\textwidth]{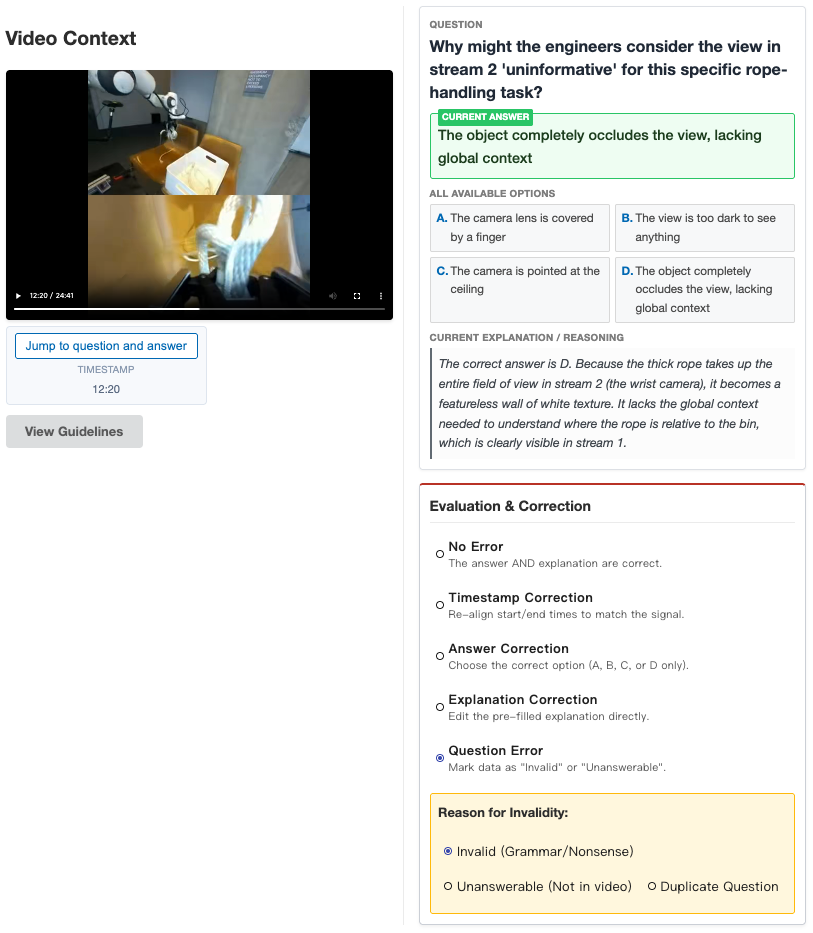}
        \caption{Question error case.}
        \label{fig:mturk2}
    \end{subfigure}
    \caption{Examples of Human annotation interfaces in MTurk. (a) if the annotator chooses timestamp correction, the correct time range should also be provided. (b) if the annotator chooses question error, the error reason range should also be provided.}
    \vspace{-1cm}
    \label{fig:mturk_examples}
\end{figure}

\subsection{Human Annotation Protocol}

In our human annotation process shown in Fig.~\ref{fig:mturk1} and Fig.~\ref{fig:mturk2}, annotators are presented with synchronized multi-stream video context, a timestamped question, a pre-filled answer, and a corresponding explanation, and are asked to verify the validity and accuracy of the annotation instance under a structured quality-control protocol. The overall standard requires workers to ground their judgment strictly in the visible video evidence at the specified temporal segment, evaluate whether the question is well-formed and answerable from the provided streams, and assess whether the proposed answer correctly matches one of the predefined options and whether the accompanying rationale is faithful to the observed scene. When the annotation is correct, workers mark it as No Error; otherwise, they identify the error type by selecting among Timestamp Correction for temporal misalignment, Answer Correction for incorrect multiple-choice selection, Explanation Correction for inadequate or inaccurate reasoning, or Question Error when the prompt is invalid, nonsensical, duplicated, or unanswerable from the video. If needed, annotators additionally provide corrected temporal boundaries or specify the reason for invalidity. Overall, the workflow follows a verification-and-correction paradigm in which workers first inspect the relevant video segment, then compare the question, answer, and explanation against the observable evidence, and finally either confirm the instance or revise its temporal, semantic, or linguistic components, thereby ensuring annotation reliability, interpretability, and consistency for downstream dataset construction and evaluation.

\begin{figure}[t]
  \centering
  \small
  \begin{minipage}[c]{0.40\linewidth} 
    \centering
    \setlength{\tabcolsep}{6pt}
    \renewcommand{\arraystretch}{1.15}
    \captionof{table}{Human correction statistics on the evaluation set.}
    \label{tab:human_correction_stats}
    \resizebox{1.0\linewidth}{!}{%
      \begin{tabular}{l|l}
        \toprule
        \rowcolor{gray!30} Metric & Value \\ \hline
        Human Correction Rate & 25.6\% \\
        Accuracy After Correction & 94.5\% \\
        \bottomrule
      \end{tabular}
    }
  \end{minipage}\hfill%
  \begin{minipage}[c]{0.48\linewidth}
    \centering
    \includegraphics[width=\linewidth]{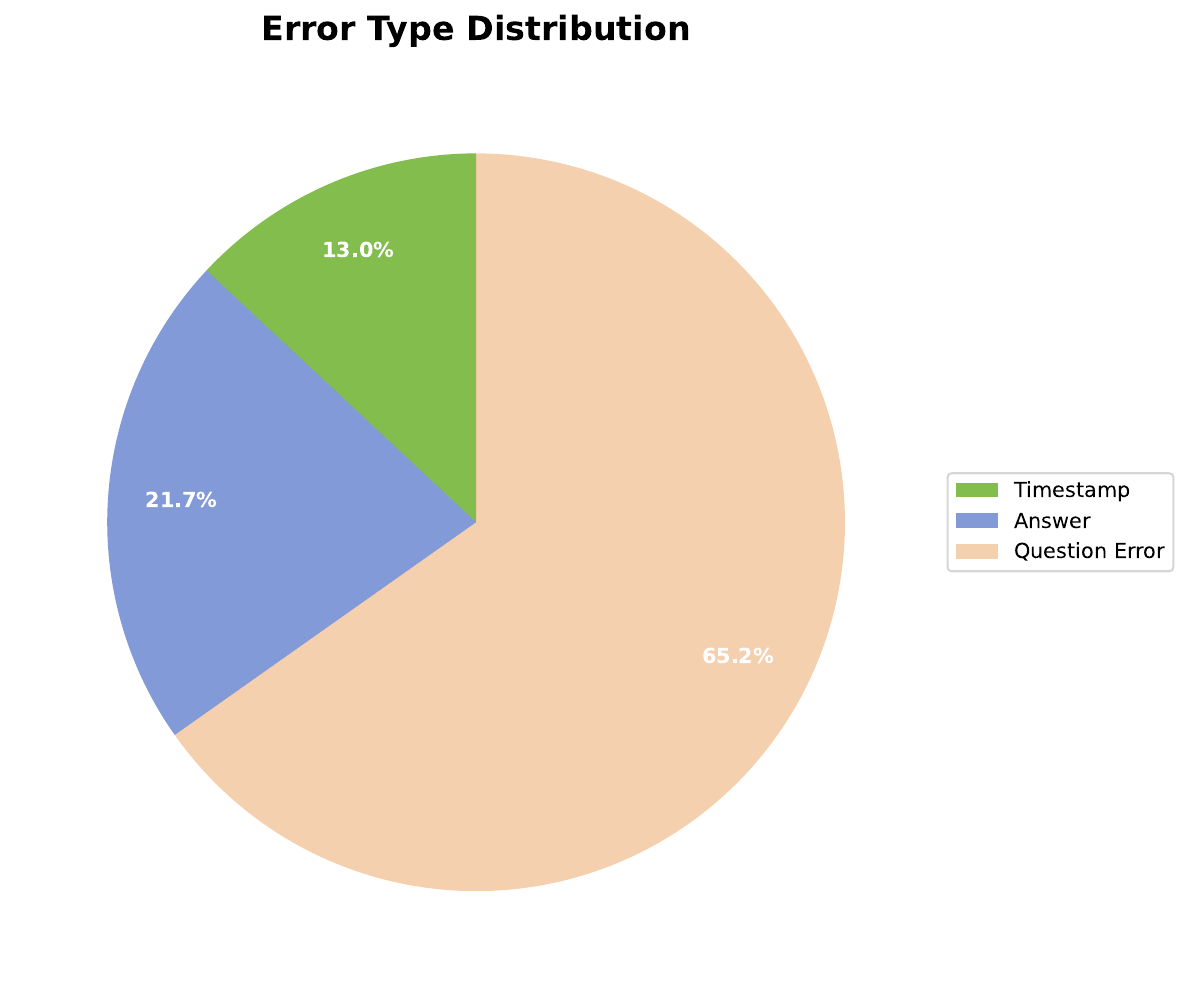}
    \captionof{figure}{Error type distribution on the human evaluation.}
    \label{fig:error_type_distribution}
  \end{minipage}
\vspace{-0.6cm}
  
\end{figure}

\vspace{-0.2cm}
\subsection{Human Correction Statistics}
\vspace{-0.2cm}

As shown in Tab.~\ref{tab:human_correction_stats}, human correction analysis on the evaluation set shows that 25.6\% of cases required manual correction, while the accuracy after correction reached 94.5\%. Among the corrected errors, question errors accounted for the largest proportion (65.2\%), followed by answer errors (21.7\%) and timestamp errors (13.0\%). These results indicate that most errors stemmed from problems in question generation or interpretation, whereas answer and timestamp issues were comparatively less frequent.

\begin{table}[!ht]
\centering
\vspace{-0.3cm}
\caption{\textbf{The licenses for the multi-domain video/data sources} used in our study.}
\label{tab:data_sources_license}
\vspace{-0.3cm}
\resizebox{0.8\textwidth}{!}{
\begin{tabular}{lll}
\toprule
\rowcolor{gray!30} \textbf{Dataset}                               & \textbf{Type}                   & \textbf{License}        \\ \hline
brain4cars~\cite{jain2016brain4cars}           & Driving                         &         BSD 2-Clause          \\
Waymo-E2E~\cite{xu2025wod}                     & Driving                         & Waymo Dataset License   \\
Apidis-Basketball~\cite{VanZandycke_DeepSport} & Sports                          &    Apidis Academic License          \\
e-Sports (Self-record)                         & Sports                          &         CC BY-NC 4.0         \\
Split-screen Game (Youtube)                    & Sports                          & CC BY-NC 4.0 / YouTube Standard License \\
DROID~\cite{khazatsky2024droid}                & Robot                           & CC BY 4.0               \\
UAV-VisLoc~\cite{xu2024uav}                & Robot                           & Apache 2.0               \\
EgoExo4D~\cite{grauman2024ego}                 & Daily Routine                   & Ego4D License           \\
EgoLife~\cite{yang2025egolife}                 & Daily Routine                   &         MIT License         \\
Seamless-interaction~\cite{agrawal2025seamless}& Chat                            & CC BY-NC 4.0            \\
WILDTRACK~\cite{chavdarovawildtrack}           & Surveillance                    &         None          \\
All-Day~\cite{fan2025all}                      & Surveillance                    &         CC BY-NC 4.0        \\
FaceEngage~\cite{chen2019faceengage}           & Live Streaming                  &         CC BY-NC 4.0 / YouTube Standard License       \\
Streamer-React (Youtube)                       & Live Streaming                  & CC BY-NC 4.0 / YouTube Standard License\\
Map-Street (Baidu/Google Map API)              & Interfaces                      & API Terms of Service    \\
Comma2K-19~\cite{schafer2018commute} $w/.$ dashboard & Interfaces                & MIT License             \\ \bottomrule
\end{tabular}
}
\vspace{-0.3cm}
\end{table}

\subsection{License and Data Usage}

We conduct a systematic review of the open-source licenses for the datasets we use, with the results summarized in Tab.~\ref{tab:data_sources_license}. The analysis indicates that CC BY 4.0 and Apache License 2.0 are the most widely adopted. After comprehensive consideration, our $X$-Stream dataset adopts the \textbf{\href{https://creativecommons.org/licenses/by/4.0/}{Creative
 Commons Attribution (CC BY) 4.0}} or \textbf{\href{https://www.apache.org/licenses/LICENSE-2.0}{Apache License 2.0}} for different sources of data, which is already used by most of the source data. 

\section{Experiment Detail}

\textbf{Token limitation:}: Tokens may slightly differ due to inherent differences in the processor. Token control is affected by the diverse visual tokenization strategies across models: \textbf{1) Gemini:} Fixed 263 tokens/sec (independent of resolution/FPS). \textbf{2) Qwen3-VL:} 28$\times$28 pixel patches per token with token merging. \textbf{3)} \textbf{GPT-5:} 85 tokens/frame + 170 tokens per 512$\times$512 tile. Since these mechanisms require different adjustments, speeding and resizing videos are necessary compromises to establish a baseline. $r_n$ and $r_n$ ($C_{max}$=$250$) depend on the pixel dimensions to ensure: 1) GPT: a maximum edge of $512$; 2) Qwen: $511$×$383$ or equivalent; 3) Gemini: irrelevant.

\section{Additional Observations and Analyses}

\subsection{Single-Stream Shortcut Cases}

The table distinguishes between shortcut cases in temporal question answering and cases that require genuine cross-stream grounding. In shortcut scenarios, such as stable global context, dominant ongoing activity, and cross-stream redundancy, the model can often answer correctly without precisely aligning events across streams, instead relying on persistent context or redundant visual cues. In contrast, genuine grounding is necessary in situations involving state changes over time, causal reactions, or transient visual attributes, where the answer depends on verifying what happens at a specific moment. Overall, the table highlights that strong performance on temporal QA does not always indicate true temporal grounding ability, as some questions can be solved through single-stream shortcuts.

\begin{table}[htbp]
    \centering
    \scriptsize
    \vspace{-0.3cm}
    \caption{Classification and Analysis of Pseudo Multi-stream QA Phenomena}
    \vspace{-0.3cm}
    \label{tab:pseudo_qa_analysis}
    \resizebox{0.9\textwidth}{!}{%
    \begin{tabularx}{\textwidth}{@{} p{0.12\textwidth} p{0.18\textwidth} >{\raggedright\arraybackslash}X >{\raggedright\arraybackslash}X @{}}
        \toprule
        \textbf{Category} & \textbf{Mechanism} & \textbf{Description} & \textbf{Example} \\
        \midrule
        
        \multirow{12}{=}{\textbf{1. Pseudo Reference}} & 
        \textbf{Invalid} \newline \textbf{Temporal} \newline \textbf{Anchoring} & 
        Occurs when the target action is continuous or static, rendering the cross-stream temporal constraint meaningless. Since the answer is invariant to time, the specific timestamp from the reference stream becomes redundant. & 
        \textit{Query:} "What is the person doing in Stream A when the door opens in Stream B?" \par
        \textit{Issue:} The person is "sitting" throughout the entire video. The specific timing of the "door opening" (anchor) is irrelevant to the answer. \\
        \cmidrule(l){2-4} 
        
        & 
        \textbf{Invalid} \newline \textbf{Visual} \newline \textbf{Anchoring} & 
        Arises when the target object in the queried stream is unique or salient enough that the visual descriptor provided by the reference stream is not required for identification. & 
        \textit{Query:} "Find the object in Stream A that matches the color of the ball in Stream B." \par
        \textit{Issue:} Stream A contains only one object. The model can identify it without needing the color information (anchor) from Stream B. \\
        \midrule
        
        \textbf{2. Pseudo Coop.} & 
        \textbf{Information} \newline \textbf{Redundancy} & 
        Happens when streams share overlapping fields of view or semantic content. The model can resolve the query using a single stream alone, bypassing the need for genuine multi-view fusion or collaboration. & 
        \textit{Query:} "Identify the object held by the person." \par
        \textit{Issue:} Due to overlapping views, the object is clearly visible in Stream A alone. The model ignores Stream B entirely. \\
        
        \bottomrule
    \end{tabularx}
    }
    \vspace{-0.2cm}
\end{table}

\subsection{Analysis of Grid Layout and Spatial Division}

During the exploratory experiment stage, we observed that different configurations of spatial division multiplexing lead to different outcomes.
Empirically, vertical-level stitching generally performs better than horizontal-level stitching.
We attribute this difference primarily to the influence of raster order on the attention mechanism. Both the earlier Qwen 2D raster order and the more recent Qwen2.5 or Qwen3 scheme, which applies 2×2 local spatial merging followed by raster-order flattening, can be viewed as variants of the classical raster scan: flattening $(T,H,W)$ in C-order, where the spatial dimensions are traversed in row-major order (left to right, then top to bottom), and the temporal dimension is traversed from earlier frames to later frames.

Under vertical-level stitching, different streams remain more clearly separable. By contrast, under horizontal-level stitching, tokens from different streams at the same time step become interleaved. This makes vertical-level stitching more conducive to forming a coherent global understanding, while reducing misinterpretation caused by token interleaving.

\begin{figure}[htbp]
\centering
\vspace{-0.2cm}
\includegraphics[width=0.8\textwidth]{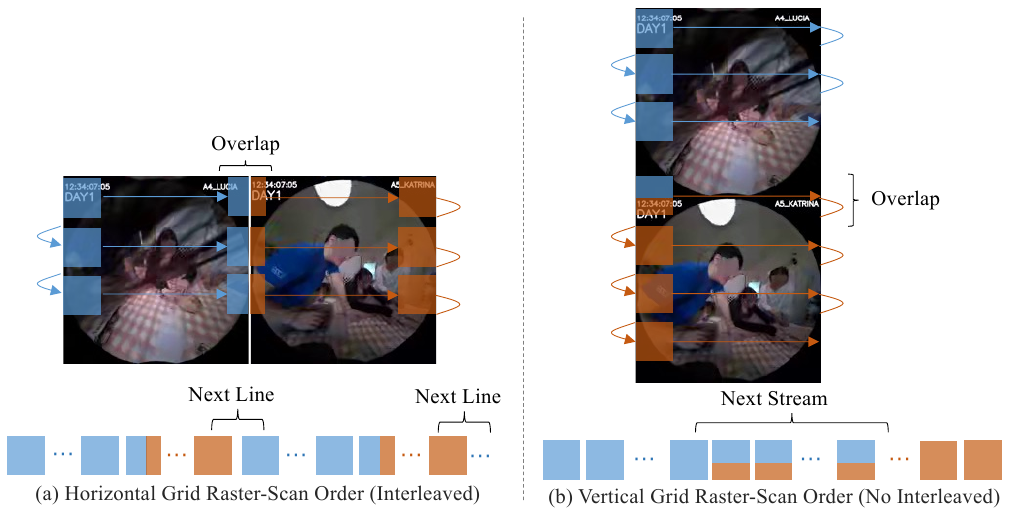}
\vspace{-0.2cm}
\caption{The difference in Grid Raster-scan order causes a performance gap. 
In general, raster-scanning can be understood as a left-to-right, top-to-bottom process.
For horizontal concatenation, scanning typically results in tokens from multiple streams being interleaved within the same frame.
Conversely, vertical concatenation generally prevents tokens from multiple streams from interleaving during scanning.
Regardless of the method used, however, a certain amount of overlapping tokens is inevitable.
}
\label{fig:grid_distribution}
\vspace{-0.2cm}
\end{figure}

\begin{table}[htbp]
\centering
\caption{Comparison of grid layout. Vertical Spatial Division always performs better than horizontal.}
\resizebox{0.4\textwidth}{!}{%
\begin{tabular}{l|cc}
\toprule
\rowcolor{gray!30} \textbf{Spatial Division} & \textbf{\begin{tabular}[c]{@{}l@{}}Qwen-3-Omni\\ -32B-A3B\end{tabular}} & \textbf{\begin{tabular}[c]{@{}l@{}}Qwen-3-VL\\ -32B-A3B\end{tabular}} \\ \hline
Vertical                                     & \textbf{34.28}                                                          & \textbf{34.19}                                                        \\
Horizontal                                   & 31.73                                                                   & 30.40                                                                 \\ \bottomrule
\end{tabular}
}
\end{table}

\vspace{-0.4cm}
\subsection{Analysis of Temporal Embedding for Time Division}
\vspace{-0.1cm}

Since Time Division is interleaved across the time dimension, only the tokens from a single stream exists at any given time step. This necessitates assigning continuous timestamps to tokens from different streams. If identical timestamps were applied, the model would be completely unable to distinguish between distinct moments, resulting in a performance degradation of up to 30\% and the loss of its core capabilities. Therefore, assigning continuous timestamps to tokens across different streams is a highly intuitive approach. Due to length constraints, the rationale of the first sample is provided.
\vspace{-0.4cm}

\section{Qualitative QA Examples and Evaluation Prompt}
\vspace{-0.1cm}

\subsection{Evaluation Prompt for LLM-as-a-Judge}
\vspace{-0.1cm}

Our prompt of LLM-as-Judge is listed below.  For consistency, the final score is rescaled to a 0–100 range. The evaluation prompt for LLM-as-a-Judge and Human Test is shown below. Note: we use Qwen3-235B-A22B for evaluation.

\vspace{-0.6cm}
\begin{figure}[htbp]
\centering
\begin{adjustbox}{width=1\textwidth}
\begin{minipage}{1.7\textwidth}
\begin{lstlisting}[label={lst:evaluator}]
You are an expert evaluator judging whether a model's answer provides a reasonable and factually plausible explanation that directly addresses the question, based on the reference answer.

**Evaluation Guideline:**
- Focus on whether the model gives a coherent reason that logically explains what the question asks.
- The answer does not need to reproduce all details from the reference - it only needs to offer a factually grounded and relevant cause.
- An answer that captures the essential reason should be considered strong, even if it omits descriptive details.
- Accept simplified, rephrased, or high-level reasoning as long as it is consistent with the reference, plausibly explains the phenomenon in the question, and does not contradict known facts.
- Do not deduct points for omitting secondary or illustrative details when the core causal logic is present, or for using concise or abstract phrasing.
- For responses in JSON or other formats, try to parse them first and then make a judgment.
- When the question is multiple-choice, also get the answer (like A, B) before making a judgment.
- Only penalize if the explanation is factually wrong, fails to provide a meaningful cause, or is so vague that it does not actually answer the question.

**Scoring (integer 0-5):**
- 5: Fully accurate and complete explanation.
- 4: Correct and logically sufficient explanation; may omit non-essential details but captures the essential reason.
- 3: Partially relevant but weakens or misses part of the core causal link.
- 2: Tangential or speculative without solid grounding.
- 1: Factually incorrect.
- 0: No attempt to answer or completely off-topic.

**Output Format:**
Return a valid JSON object with exactly two keys:
- "explanation": one sentence focusing on whether the answer gives a reasonable and relevant reason for the question
- "score": an integer from 0 to 5

Output only the JSON. No other text, markdown, or commentary.

**Inputs:**
- Question: {question}
- Predicted Answer: {model_output}
- Correct Answer: {reference_answer}
\end{lstlisting}
\end{minipage}%
\end{adjustbox}
\vspace{-1.4cm}
\end{figure}

\end{document}